%% file: main.tex
\documentclass{article} 
\usepackage{iclr2027_conference,times}

\input{math_commands.tex}

\usepackage{amssymb}
\usepackage{booktabs}
\usepackage{multirow}
\usepackage{longtable}
\usepackage{array}
\usepackage{float}
\usepackage{flafter}

\usepackage{graphicx}
\usepackage{url}
\usepackage{hyperref}
\hypersetup{hidelinks}
\setcitestyle{citesep={,}}
\definecolor{revisionblue}{gray}{0}
\newenvironment{revision}{\begingroup}{\par\endgroup}

\title{Where Structure Enters a World Model: From Predictive Fit to Physical Rollout}

\author{Panayiotis Christou \\ Fordham University
        \And Satyapragnya Kar \\ New York University
        \And Flavjo Xhelollari \\ Fordham University
        \AND Parijat Dube \\ Oracle
        \And Rajarshi Das \\ MQube Cognition}
\hypersetup{pdftitle={Where Structure Enters a World Model: From Predictive Fit to Physical Rollout},
            pdfauthor={Panayiotis Christou, Satyapragnya Kar, Flavjo Xhelollari, Parijat Dube, Rajarshi Das}}
\iclrfinalcopy
\begin{document}

\maketitle
\lhead{Preprint}

\input{paper_sections/01_abstract}
\input{paper_sections/02_introduction}
\input{paper_sections/03_background}
\input{paper_sections/04_related_work}

\input{paper_sections/05_problem_formulation}
\input{paper_sections/06_solution_design}

\input{paper_sections/07_validation}
\input{paper_sections/08_results}

\input{paper_sections/09_discussion}

\input{paper_sections/10_conclusion}

\label{maintext:end}
\input{paper_sections/12_statements}
\bibliographystyle{iclr2027_conference}
\bibliography{references}

\appendix
\input{paper_sections/11_appendix}

\input{paper_sections/13_revision_controls}

\end{document}

%% file: math_commands.tex
\usepackage{amsmath,amsfonts,bm}

\def\eqref#1{equation~\ref{#1}}

\def\1{\bm{1}}

\DeclareMathAlphabet{\mathsfit}{\encodingdefault}{\sfdefault}{m}{sl}
\SetMathAlphabet{\mathsfit}{bold}{\encodingdefault}{\sfdefault}{bx}{n}



%% file: paper_sections/01_abstract.tex
\begin{abstract}
Predictive world models learn to forecast observations, but accurate short-term
predictions do not necessarily produce physically coherent behavior over longer
horizons. We investigate how incorporating physical structure can strengthen
world models while retaining their underlying predictive objectives. Across
token-based models of planetary motion and JEPA-style visual models of
mechanical systems, we study conserved-quantity supervision, symmetry-aware
representations, temporal consistency, and learned structured dynamics.
In planetary motion, angular-momentum supervision improves autonomous
next-token forecasting, while the benefits of temporal regularization depend
on the accompanying physical objective and representation. In visual models,
generating rollouts through supervised energy-based dynamics heads produces
more faithful trajectories and phase-space evolution. Together, these findings
highlight the importance of how physical structure participates in prediction:
improvements in an auxiliary physical readout or a local acceleration estimate
do not automatically translate into more reliable long-horizon behavior.
Evaluations across initial conditions and physical parameter shifts further
reveal where these benefits persist and where generalization remains limited.
Our study supports physical structure as a practical means of improving
predictive world models and provides a unified framework for understanding its
effects on representations, learned transitions, and autonomous rollouts. The
central lesson is that effective physical inductive biases should help models
both represent physical regularities and use them to generate future states.
\end{abstract}

%% file: paper_sections/02_introduction.tex
\section{Introduction}
\label{sec:introduction}

World models used for planning must predict what happens after observations
stop. Small errors can accumulate until a pendulum loses its oscillation or
a planet leaves its orbit. Can physical structure help a predictive model
sustain the motion of a system while relying on its own predictions?

We add physical structure in four ways: supervising physical quantities,
building relative geometry into the model, encouraging consistent representation
changes over time, and learning mechanical equations for forecasting.
These choices guide training or forecast generation. Prior work examines
what models represent \citep{li2023othello,mitchell2023understanding} and how
consistently they behave \citep{vafa2024worldmodel}. We ask how physical
guidance can improve the motion they generate.

\begin{revision}
We study two model families. A joint-embedding predictive architecture (JEPA)
infers motion from images, progressing from oscillating pendulums and springs
to damping and coupled double-pendulum motion. For planetary motion, a
separate autoregressive transformer receives positions directly and generates
coordinates through next-token prediction (NTP). Together, these studies
test physical guidance when the model must infer state and when state is observed.
Attribution comes from matched comparisons within each family, rather than
from comparing the two modalities or architectures directly.
\end{revision}

These four forms of structure lead to three main findings:\nopagebreak
\begin{itemize}
    \item \textbf{Physical guidance improves generated trajectories.}
    Learned mechanics improves visual rollouts across four systems.
    Angular-momentum supervision improves planetary forecasts even though
    its auxiliary head is unused during generation.
    \item \textbf{Geometry improves forecasting within NTP.}
    A pairwise adapter improves the matched relative-coordinate model while
    retaining the next-token objective and decoder.
    \item \textbf{Temporal benefits depend on the representation and task.}
    Controlled comparisons identify gains in forecasting and force learning,
    including force-direction transfer under eccentricity and stellar-mass changes.
\end{itemize}

%% file: paper_sections/03_background.tex
\section{Background and Related Work}
\label{sec:background}

\paragraph{Learning what changes over time.}
JEPA learns to predict features of observations, with V-JEPA extending this
objective to masked video \citep{assran2023ijepa,bardes2024vjepa}.
Temporal objectives guide how those features change. NextLat predicts the
next latent, while Temporal Difference in Vision (TDV) uses observed image
changes to predict representation changes \citep{teoh2025nextlat,daithankar2026tdv}.
We adapt TDV to planetary motion. Related work introduces spatial smoothness,
stability, and temporal locality into GPT-2-scale models through continuous
coordinate regression, noisy contexts, and restricted attention
\citep{liu2026keplernewton}. Our models retain discrete NTP and its causal context.

\paragraph{Using physics to guide prediction.}
Physical information can shape a learning signal or the rule that advances
state. Equation-residual penalties \citep{raissi2019pinn}, differentiable
solvers \citep{hosseini2025phirom}, and physical controls
\citep{gillman2025forceprompting} offer different ways to make this connection.
Hamiltonian and Lagrangian networks learn functions such as energy and derive
motion through their derivatives \citep{greydanus2019hamiltonian,cranmer2020lagrangian}.
Dissipative extensions capture energy loss \citep{sosanya2022dissipative}.
Context-sensitive imagination \citep{roder2025dali} and physical-state recovery
from predicted features \citep{rao2026skyjepa} also link physical structure to
forecasting. We compare physical targets used during NTP training with
supervised mechanics used to generate visual rollouts.

%% file: paper_sections/04_related_work.tex
\begin{revision}
\paragraph{Making geometry part of the model.}
For a planar position \(r_i\), a rigid transform is \(r_i'=Qr_i+a\),
with translation \(a\in\mathbb R^2\) and orthogonal matrix \(Q\in O(2)\).
The set \(O(2)\) contains all planar rotations and reflections.
Invariant scalars stay unchanged. Equivariant vectors rotate or reflect
with their inputs. Relative coordinates remove shared translation, while
summed messages remove dependence on source ordering \citep{zaheer2017deepsets}.
Weighting relative vectors with invariant scalars gives equivariant updates
\citep{satorras2021egnn}. This construction connects object interactions
\citep{lei2025spartan,feng2025fioc} and temporal differences
\citep{yuan2025nsegnn} to learned dynamics. We use it to propose displacements
inside NTP, where the useful symmetries remain task-dependent
\citep{vadgama2025probing}.
\end{revision}

Conservation means constancy along a trajectory. Signed planar angular
momentum, for example, is conserved in our orbits yet reverses under
reflection. We therefore test conservation supervision and geometry separately.

\begin{revision}
\paragraph{Testing what the model can do.}
Inductive-bias probes assess behavior after controlled adaptation
\citep{vafa2025inductivebias}. We pair that perspective with autonomous
forecasting. Forecasting tests generated motion, force adaptation tests
learning from limited labels, and force transfer tests reuse on changed orbits.
\end{revision}

\begin{revision}
\paragraph{Separating structure from supervision.}
\citet{botev2021priors} find that stronger physical priors do not consistently
improve latent dynamics. \citet{gruver2022deconstructing} show that coordinates
and acceleration modeling can explain apparent Hamiltonian advantages.
Our shared second-order dynamics and RK4 control tests energy parameterization
beyond those factors. Its system- and metric-dependent gains accord with
Botev's findings. Image-based mechanics learns through reconstruction
\citep{toth2020hgn,zhong2020lagcavae}, whereas our decoder is state-supervised.
Noether Networks learn conservation within prediction \citep{alet2021noether},
and graph simulators propagate particle states \citep{sanchezgonzalez2020gns}.
Our contribution is controlled evidence separating transition accuracy,
conservation and geometric robustness in predictive models.
Appendix~\ref{app:revision_controls} includes a Lag-caVAE comparison.
\end{revision}

%% file: paper_sections/05_problem_formulation.tex
\section{Problem Setting}
\label{sec:problem_formulation}

\begin{revision}
Both tasks initialize from observed history, then forecast without further
observations or simulator access. Their models and objectives are separate.
We ask:
\textbf{RQ1:} How does learned mechanics affect autonomous visual forecasts?
\textbf{RQ2:} Can physical supervision and pairwise geometry improve forecasting
while retaining NTP?
\textbf{RQ3:} How does temporal supervision affect planetary forecasting and
force learning across different model representations?
\textbf{RQ4:} Do the improvements persist when physical parameters or forecast
horizons change?
\end{revision}

\begin{revision}
\paragraph{Visual forecasting.}
Given two images \(x_{t-1},x_t\), the model must predict future physical
states \(y_{t+k}=(q_{t+k},v_{t+k})\), where \(t\) indexes frames and \(k\)
counts forecast steps. Here \(q\) is an angle or spring position and \(v\)
its velocity. Encoder \(f_\theta\) and predictor \(g_\phi\) predict next-image
features from the observed pair \(x_{\leq t}\). During training, target encoder
\(\bar f_{\bar\theta}\) processes the next image together with that history.
Subscripts \(\theta,\phi,\bar\theta\) denote network parameters.
Mask \(m\) selects the patch positions to compare:
\begin{equation}
 \mathcal L_{\mathrm{vid}}
 =\ell\!\left(g_\phi(f_\theta(x_{\leq t}),m),
       \operatorname{sg}[\bar f_{\bar\theta}(x_{t+1}\mid x_{\leq t})_m]\right),
       \label{eq:video_base_loss}
\end{equation}
Here \(\ell\) is Smooth-L1 loss on normalized features. Subscript \(m\)
selects target patch features. The stop-gradient operator
\(\operatorname{sg}\) leaves target values unchanged but blocks backpropagation
through the target branch. Target weights follow an
exponential moving average of the trained network parameters. Both visual models
also use numerical position and velocity labels to train the encoder,
predictor, and state decoder. Physics-JEPA additionally trains a mechanical
transition (Section~\ref{sec:visual_method}).
At inference, both models receive only the initial image pair.
\end{revision}

\begin{revision}
\paragraph{Planetary forecasting.}
A separate transformer learns from sequences of \(N\) discretized coordinate rows.
After a sampling-cadence header, each row \(z_i\) contains one body's
coordinates. Rows enumerate bodies within a frame, then advance in time.
The learned distribution \(p_\theta\) predicts the next row from the
preceding rows \(z_{\leq i}\):
\begin{equation*}
 \mathcal L_{\mathrm{NTP}}
 =-\sum_{i=1}^{N-1}\log p_\theta(z_{i+1}\mid z_{\leq i}).
\end{equation*}
Let \(\mathcal K\subseteq\{F,H,L,\mathrm{TD}\}\) index the force, orbital-energy,
angular-momentum, and temporal objectives enabled in a planetary variant.
With weights \(\lambda_k\geq0\), its training loss is
\begin{equation}
 \mathcal L_{\mathrm{planet}}=\mathcal L_{\mathrm{NTP}}
       +\sum_{k\in\mathcal K}\lambda_k\mathcal L_k.
 \label{eq:total_loss}
\end{equation}
Physical labels are calculated from simulated states. The temporal objective
uses coordinate displacements. All variants generate future coordinates
through NTP, with auxiliary heads used only in training.
Table~\ref{tab:bias_taxonomy} places these objectives alongside structure
within the model's computation.
\end{revision}

\begin{table}[tbp]
\caption{Physical structure organized by its role in predictive computation.
The remaining interaction functions and potential dynamics are learned.}
\label{tab:bias_taxonomy}
\centering
\setlength{\tabcolsep}{4pt}
\begin{tabular}{>{\raggedright\arraybackslash}p{.19\linewidth}>{\raggedright\arraybackslash}p{.20\linewidth}>{\raggedright\arraybackslash}p{.29\linewidth}>{\raggedright\arraybackslash}p{.23\linewidth}}
\toprule
Structure & Entry point & Supplied information & Capability tested \\
\midrule
Physical supervision & Training objective & Force or conserved-quantity labels & Autonomous token forecasting \\
Temporal consistency & Representation objective & Same-body displacement or a predicted physical step & Temporal evolution \\
Relative and pairwise geometry & Input and residual branch & Relative vectors and invariant scalar features & Token rollout and frame behavior \\
Learned mechanics & Executed transition & Kinetic structure and state/acceleration supervision & Visual trajectories and phase space \\
\bottomrule
\end{tabular}
\end{table}

%% file: paper_sections/06_solution_design.tex
\section{Structure within Predictive World Models}
\label{sec:method}

\begin{revision}
\begin{figure}[!tbp]
\color{revisionblue}
\centering
\includegraphics[width=\linewidth]{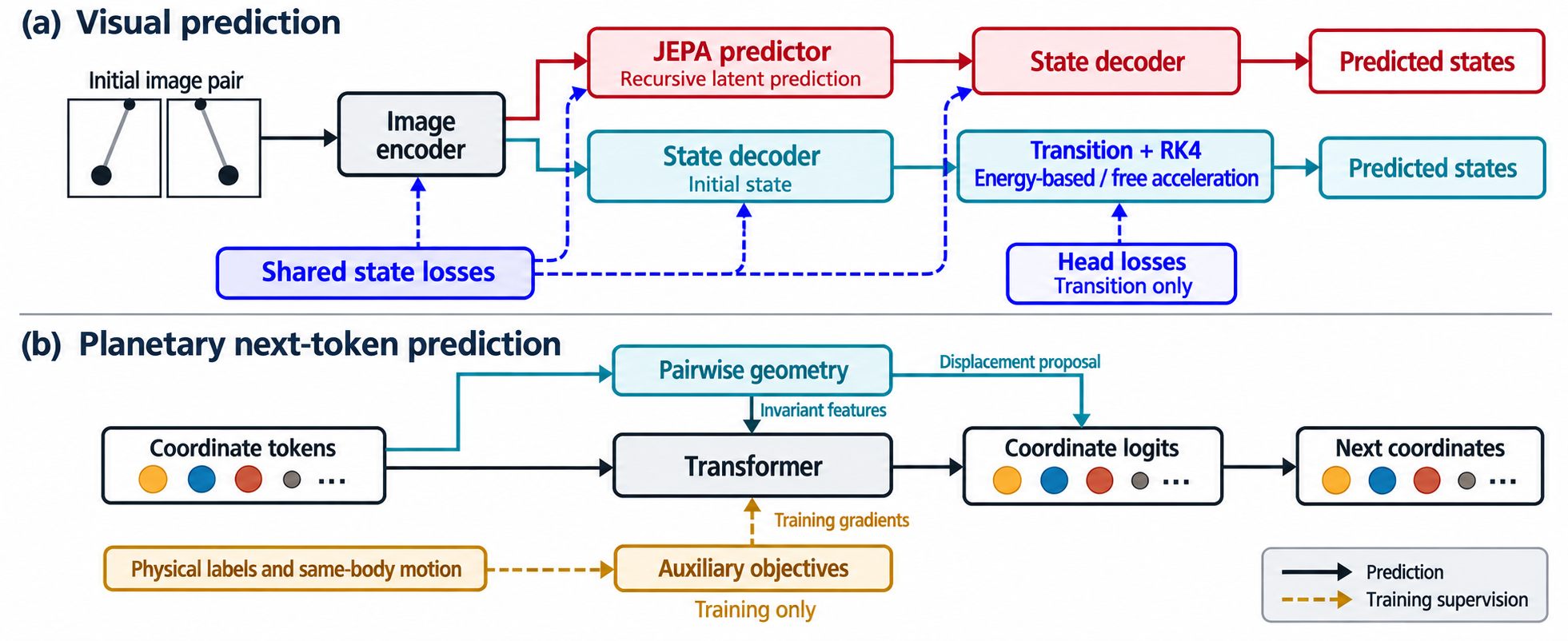}
\caption{Visual (a) and planetary (b) forecasting.
Our V-JEPA 2 baseline decodes predicted features. Physics-JEPA advances states
through learned mechanics with fourth-order Runge--Kutta integration (RK4).
Both visual models share initial images and state supervision, which updates
the encoder, predictor and decoder. Mechanics losses update only the head.
The planetary adapter changes embeddings and token scores, while auxiliary
losses train the backbone. Solid arrows show prediction, dashed arrows supervision.
Blue arrows show the state-loss paths and the matched free-acceleration
transition, which retains the decoded initial state and RK4.}
\label{fig:prediction_paths}
\end{figure}
\end{revision}

\begin{revision}
Hats denote predicted quantities throughout this section.
\end{revision}

\subsection{Visual prediction with learned mechanical transitions}
\label{sec:visual_method}

Both visual models in Figure~\ref{fig:prediction_paths} start from the same
initial image pair. Baseline V-JEPA 2 repeatedly predicts visual features
and decodes them into position and velocity. Physics-JEPA decodes the initial
state once, then advances it with learned mechanics.

\begin{revision}
Both train from scratch using the visual feature loss in
Equation~\ref{eq:video_base_loss} and mean squared errors on decoded
positions and velocities, for both observed and predicted frames.
The mechanical head learns a potential energy \(V_\psi(q)\) and a
dissipation function \(R_\psi(q,v)\), where \(\psi\) denotes its parameters.
Dots denote time derivatives and \(\partial_q,\partial_v\) derivatives with
respect to position and velocity. For a single angle or spring coordinate,
energy \(E_\psi\) uses fixed kinetic energy with unit inertia:
\begin{equation*}
 E_\psi(q,v)=\tfrac12v^2+V_\psi(q),\qquad
 \dot q=v,\qquad
 \dot v=-\partial_q V_\psi(q)-\partial_v R_\psi(q,v).
\end{equation*}
The potential's slope determines acceleration.
Conservative systems set \(R_\psi=0\). Damped systems learn
a positive damping coefficient \(d_\psi(q)\) and use
\(R_\psi(q,v)=\tfrac12d_\psi(q)v^2\), so that
\(\dot E_\psi=-d_\psi(q)v^2\).
For the double pendulum, \(q\) contains both angles and \(v\) both angular
velocities. The head uses their known coupled kinetic energy
(Appendix~\ref{app:mechanics}).
RK4 integrates these equations over each frame interval.
Physics-JEPA trains the head on simulated states, penalizing errors in
acceleration and energy balance along the motion. These losses
update only the mechanics head. Separate gradient clipping preserves the
visual updates. A second, Lagrangian head receives acceleration supervision
during training (Appendix~\ref{app:mechanics}).
The baseline uses only feature and state losses.
In the added matched controls, frozen observation weights feed energy and
free-acceleration heads with the same labels, shared-layer initialization,
budget and RK4. Both use $\dot q=v$. Acceleration-only, next-state and decoded
transition objectives exclude the original energy-balance penalty.
Appendix~\ref{app:background_equations} tests an additional loss that couples
mechanics to visual prediction.
\end{revision}

\subsection{Physical supervision while retaining next-token prediction}

\begin{revision}
For planetary models, auxiliary readouts predict force \(F\), orbital energy
\(H\), and planar angular momentum \(L_z\) from each body's hidden state.
Their mean squared error (MSE) losses are
\begin{equation*}
 \mathcal L_F=\operatorname{MSE}(\hat F,F),\quad
 \mathcal L_H=\operatorname{MSE}(\hat H,H),\quad
 \mathcal L_L=\operatorname{MSE}(\hat L_z,L_z).
\end{equation*}
Let \(r_i,r_\star\) be absolute planet and star positions, \(m_i,m_\star\)
their masses, and \(G\) the gravitational constant. Newtonian force is
\begin{equation*}
 F_i=-Gm_i m_\star
 \frac{r_i-r_\star}{\lVert r_i-r_\star\rVert_2^3}.
\end{equation*}
Here, \(\|\cdot\|_2\) denotes the Euclidean norm. The cubic denominator
normalizes the displacement vector and gives an inverse-square force magnitude.
For star-relative position \(r=r_i-r_\star\), relative velocity \(v\),
and reduced mass \(\mu=m_i m_\star/(m_i+m_\star)\), the remaining targets are
\(H=\tfrac12\mu\|v\|^2-Gm_i m_\star/\|r\|\) and
\(L_z=\mu(r_xv_y-r_yv_x)\).
Labels match hidden states by body and time.
Each enabled loss in Equation~\ref{eq:total_loss} trains both readout
and backbone. NTP alone generates coordinates.
\end{revision}

\subsection{Relative coordinates and symmetry-aware interactions}

Plain NTP predicts absolute coordinates. Relative NTP predicts changes in
a body's position relative to its star. Training and generation both
compute and accumulate these displacements in integer coordinate bins.

\begin{revision}
Pairwise NTP estimates a body's next displacement from where other bodies
are and how they have recently moved. The adapter uses the latest two
complete prefix frames and makes this proposal rotate with the system.
For body \(i\) at frame \(t\),
\(\tilde r_{i,t}\) is its position divided by 50 astronomical units (AU).
Its frame displacement is
\(\delta r_{i,t}=\tilde r_{i,t}-\tilde r_{i,t-1}\), without division by timestep.
For another body \(j\), \(r_{ij}=\tilde r_{j,t}-\tilde r_{i,t}\) is their
separation and \(w_{ij}=\delta r_{j,t}-\delta r_{i,t}\) their relative motion.
Their squared lengths and dot product form three scalar features \(u_{ij}\).
A shared network with parameters \(\psi\) maps these features to a vector
message \(\gamma_\psi\) and scalar weights \(\alpha_\psi,\beta_\psi\).
Summing over other bodies gives hidden message \(\eta_i\) and displacement proposal \(d_i\):
\begin{align*}
 u_{ij}&=[\|r_{ij}\|^2,\|w_{ij}\|^2,r_{ij}^{\mathsf T}w_{ij}],
 &\eta_i&=\sum_{j\ne i}\gamma_\psi(u_{ij}),\\
 d_i&=\sum_{j\ne i}
    \alpha_\psi(u_{ij})r_{ij}+\beta_\psi(u_{ij})w_{ij}.
\end{align*}
The adapter adds \(\eta_i\) to the input embedding and biases output token
scores toward the proposed displacement \(50d_i\) AU.
One learned gate controls both additions, starting at zero.
Both learn through the NTP loss, retaining the full causal history.
Distances and dot products are rotation- and reflection-invariant, making
the continuous proposal \(d_i\) equivariant (Appendix~\ref{app:equivariance}).
Ordered attention and quantization remain in the full token model.
\end{revision}

\subsection{Same-body temporal learning}

\begin{revision}
Neighboring token rows can describe different bodies. We instead pair the
same body across successive frames. Our adaptation of TDV
\citep{daithankar2026tdv} uses coordinate displacement to predict how its
representation changes, replacing the original image differences.
Let \(h_{b,t}\) be the transformer's hidden state for body \(b\) at frame \(t\),
and \(\xi_{b,t}\) its quantized star-relative coordinates.
A small network \(M_{\mathrm{TD}}\) predicts a feature change from the current
hidden state and scaled displacement
\(\delta\xi_{b,t}=(\xi_{b,t+1}-\xi_{b,t})/c_\xi\):
\begin{equation*}
 \mathcal L_{\mathrm{TD}}
 =\rho\!\left(h_{b,t}+M_{\mathrm{TD}}
       (h_{b,t},\delta\xi_{b,t}),
       \operatorname{sg}[h_{b,t+1}]\right).
\end{equation*}
Here \(\rho\) is Smooth-L1 loss, averaged over valid same-body pairs and
hidden dimensions. The scale \(c_\xi=7002\) is the maximum token ID.
The target is the same transformer's next hidden state, with gradients stopped.
Original TDV uses an exponential-moving-average teacher.
The loss trains the temporal network and NTP backbone without force or mass labels.
Future displacement is training-only, leaving NTP generation unchanged.
Appendix~\ref{app:training_contracts} gives the settings.
\end{revision}

%% file: paper_sections/07_validation.tex
\section{Experimental Design}
\label{sec:experimental_protocol}

\begin{revision}
\paragraph{Visual training and evaluation.}
Each visual system trains on 200 simulated episodes of 150 images
(\(64\times64\) pixels), sampled every \(\Delta t=0.05\) s.
Each training window contains two observed frames and the next image as target.
The encoder uses four transformer blocks, eight heads, width 256, and
\(8\times8\) patches. We evaluate the final epoch: 50 for simple pendulum
and spring, 60 for damped pendulum, and 120 for double pendulum.
The original simple and damped pendulum comparisons have three independently initialized training
runs (random seeds 42--44), and spring and double pendulum one (seed 42).
\end{revision}

Root mean squared error (RMSE) pools 400 future steps (20 s) over
25 held-out initial conditions, excluding initialization.
Angular errors take the shortest difference around the circle.
Energy RMSE uses the simulator's energy function. Acceleration RMSE compares
velocity changes per unit time. We report means and sample standard deviations
across training runs. Appendix~\ref{app:force_metrics} defines drift metrics.

\begin{revision}
\paragraph{Planetary training and data.}
All variants share 100,000 single-planet trajectories and 15,625 AdamW
updates at learning rate \(6\times10^{-4}\).
Each update combines gradients from four batches of eight trajectories
(effective batch size 32). The transformer has 12 layers,
12 attention heads, width 768,
and 999 body-coordinate rows of context. Coordinates use 7,000
bins over \([-50,50]\) AU plus three special tokens. Relative and pairwise
models share integer-bin encoding. Auxiliary weights are 0.1.
Velocities and masses are saved in physical-state records alongside
coordinates to compute targets and evaluation metrics. They are not input tokens.
Validation NTP loss selects the checkpoint. The original three independent training runs
use random seeds 1337--1339. The adapter increases parameter count
from 108.53M to 109.72M.
Absolute TDV uses the earlier auxiliary coordinate encoding, so its ranking
against corrected relative TDV compares complete configurations.
Relative-model contrasts hold encoding fixed.
\end{revision}

Each system draws two independent masses uniformly between 0.0001 and
0.0101 solar masses, designating the heavier body as its star.
Orbit size (semimajor axis) is uniform between 0.3 and 32 AU.
Elongation (eccentricity) follows a beta distribution with shape parameters
0.867 and 3.03. The cadence token specifies regular sampling every
0.0189 or 0.486 years, shared by both bodies.

\paragraph{Planetary forecasting and physics.}
Each model receives 32 frames and generates 128 on 16 held-out systems,
choosing the most probable valid coordinate token at each step.
Position MSE averages squared errors over future planet coordinates,
then systems, excluding the star. Angular-momentum drift measures change
from the last observed state, divided by a fixed physical scale from the
true initial orbit. Equation~\ref{eq:angular_momentum_drift} defines this
scale, including near-zero angular momentum.
Finite differences estimate velocities from predicted positions.
Table~\ref{tab:planetary_main} aggregates training runs.
Appendix~\ref{app:planetary} gives individual scores and paired uncertainty intervals.

\paragraph{Sparse force adaptation.}
Each backbone receives a fresh linear force head and 1,000 full-backbone
updates using 25 force labels per orbit, 50 for validation, and 424 for
testing. Fits use eight named-planet systems at both cadences, with identical
label splits and seed 1337. Figure~\ref{fig:matched_force} shows Earth at the
finer cadence, with each condition fitted separately.
Figure~\ref{fig:generalization}(a,b) averages all 16 orbits and reuses the
nominal fit under both parameter changes without refitting. Appendices~\ref{app:matched_force_adaptation}
and~\ref{app:force_alignment_transfer} report all nine variants.
Observed orbital positions are inputs. Force-vector NMSE divides mean squared
vector error by mean squared true-force magnitude.

\begin{revision}
\paragraph{Additional matched controls.}
Appendix~\ref{app:revision_controls} extends visual controls to seeds 42--46
and planetary controls to 1337--1341. Decoded heads retain state-supervised
observation models. Equal-input planetary controls share observed-prefix
velocities and masses. These additions retain all seeds and remain separate
from the original protocols.
\end{revision}

\paragraph{Controlled changes.}
Paired planetary test sets change one parameter, retaining other random draws.
Eccentricities 0.45 and 0.75 lie within training support. Tests span 64 frames,
with a separate 400-frame test set. Further tests vary stellar mass, body count,
order, and coordinates. Multi-planet systems combine independent star-planet
orbits. Visual tests vary initial conditions, rendering, lengths, and masses.
Changed parameters are withheld from the head. Forecasts use pretrained
checkpoints without adaptation. Appendix~\ref{app:generalization} gives all scenarios.

%% file: paper_sections/08_results.tex
\section{Results}
\label{sec:results}

\subsection{Learned mechanics improves visual dynamics}
\label{sec:visual_results}

Physics-JEPA improves all four systems over the V-JEPA 2 baseline
(Table~\ref{tab:visual_main}). Both are trained to predict one step ahead
and evaluated over 400 steps from the same image pair.
Simple-pendulum angle RMSE falls from \(0.8814\) to \(0.2212\) rad,
and damped-pendulum RMSE from \(1.1261\) to \(0.0446\) rad, improving
in every seed.

\input{paper_sections/generated/visual_main}

\begin{figure}[!htb]
\centering
\includegraphics[width=\linewidth]{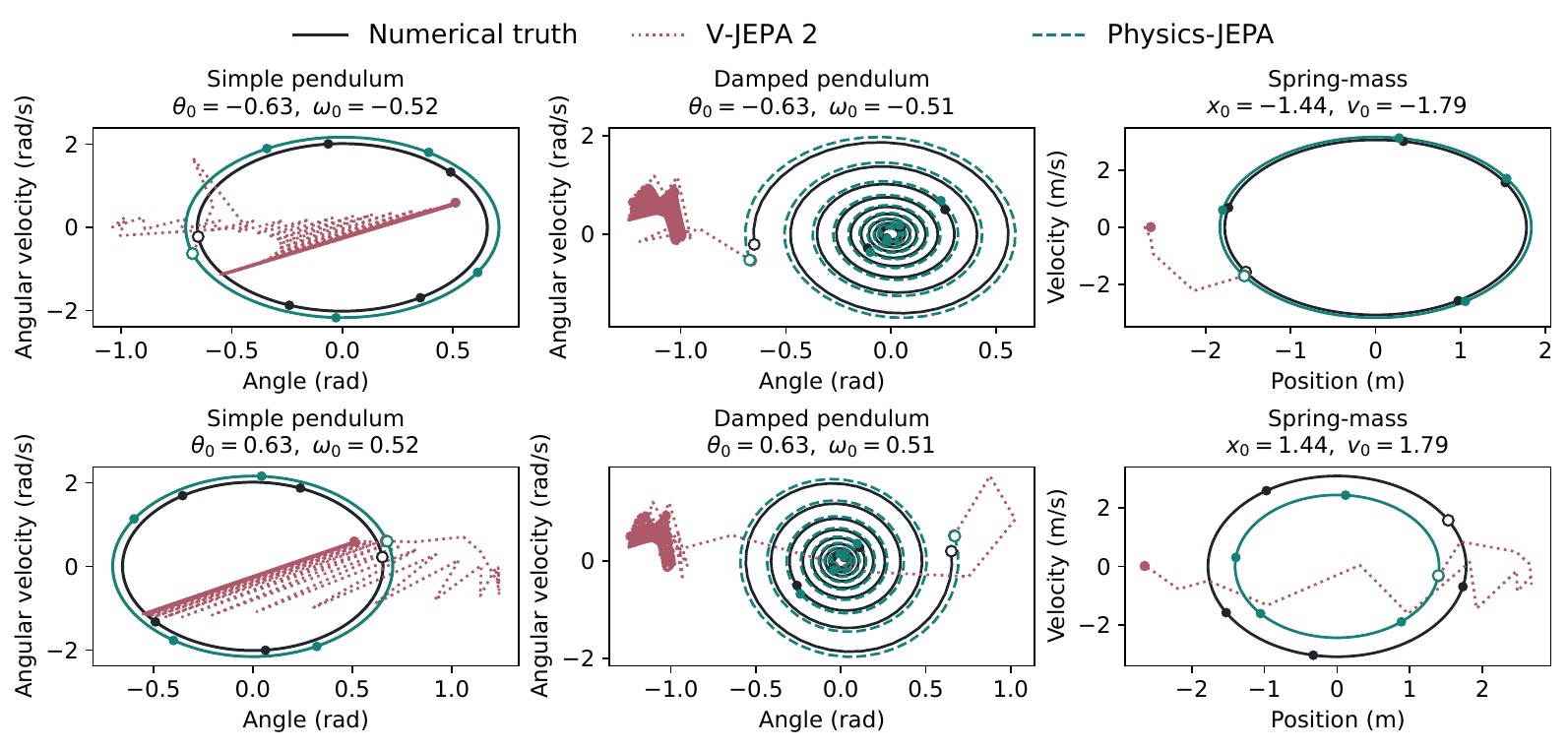}
\caption{Autonomous forecasts from the same images. The separately trained V-JEPA 2 baseline
(red) predicts visual features, while Physics-JEPA (green) follows learned
mechanics. Black curves show truth, open circles the first forecast, and dots
5 s intervals. Titles give the last observed states for two conditions and seed 42.
Aggregate scores include all 25 conditions and 400 forecast steps.}
\label{fig:visual_phase_space}
\end{figure}

In phase space, sustained oscillation traces a closed loop, while damping
draws the trajectory inward as energy is lost.
Figure~\ref{fig:visual_phase_space} shows the V-JEPA 2 baseline losing these
patterns. Learned mechanics more closely preserves the pendulum and spring's
cyclic motion and the damped pendulum's inward spiral.
Simple-pendulum acceleration RMSE falls from \(33.9603\) to \(1.8284\)
rad/s\(^2\), and energy-value RMSE improves across all systems (Table~\ref{tab:visual_main}).

\begin{revision}
Matched baseline and Physics-JEPA models produce identical latent forecasts.
Physics-JEPA improves trajectories by using its supervised mechanics head
to advance the decoded state. Latent agreement holds across eight nominal pairs and
116 parameter tests.
Appendix~\ref{app:visual} compares fresh-image prediction with autonomous forecasts.
\end{revision}

\begin{revision}
\paragraph{Matched dynamics isolate the transition.}
With matched information and RK4 (Table~\ref{tab:revision_visual}), simple-pendulum
angle error improves in all five pairs under each objective, including 18.2\%
with decoded targets.
Damped next-state heads have nearly equal angle error, but energy-balance
error is 5.9 times lower with energy dynamics. Spring drift is 9.93/6.44 times
lower under acceleration/next-state training, with position differences below
0.03\%. Double-pendulum angle errors improve by about 1\%/3.6\% in every pair
for both links, while next-state energy-change error favors free dynamics
(0.08831 versus 0.13381). Accuracy and consistency are distinct benefits.
\end{revision}

\begin{revision}
\begin{table}[tbp]
\color{revisionblue}
\caption{Matched simple-pendulum controls, five-seed means over 400 steps.
Angles are radians and physical energy drift uses simulator units.
Appendix~\ref{app:revision_visual} reports seed variation and all four systems.}
\label{tab:revision_visual}
\centering
\begin{tabular}{lrrrr}
\toprule
Head targets & \multicolumn{2}{c}{Angle RMSE} & \multicolumn{2}{c}{Energy-drift RMSE}\\
 & Energy & Free & Energy & Free\\
\midrule
Acceleration & 0.2147 & 0.2477 & 0.000150 & 0.005856\\
Next state & 0.2149 & 0.2574 & 0.000145 & 0.006771\\
Decoded transitions & 0.5117 & 0.6253 & 0.07246 & 4.27361\\
\bottomrule
\end{tabular}
\end{table}
\end{revision}

\subsection{Physical supervision and geometry improve token forecasting}
\label{sec:planetary_results}

Can structure also improve the existing predictor? Planetary motion tests
this question with NTP generating every future coordinate.

\begin{revision}
\paragraph{Angular-momentum supervision improves autonomous NTP.}
Plain NTP's position MSE falls from \(1.0755\) to \(0.1437\) AU\(^2\),
with lower angular-momentum drift (Table~\ref{tab:planetary_main}).
Both improve in every original matched seed. The auxiliary head is unused during
generation, so the token predictor itself improves.
\end{revision}

\input{paper_sections/generated/planetary_main}

\begin{revision}
\paragraph{Pairwise geometry improves relative NTP.}
Pairwise NTP reduces relative NTP's position MSE from \(0.2034\) to \(0.0630\)
AU\(^2\) (Table~\ref{tab:planetary_main}). Eight-frame displacement-vector
normalized mean squared error (NMSE) falls from \(0.1155\) to \(0.0301\)
(Table~\ref{tab:planetary_motion_vectors}). NMSE scales error by the true vectors'
mean squared magnitude. Both improve in every original seed with the same coordinates
and token encoding.
\end{revision}

\begin{revision}
\paragraph{Capacity and information controls.}
Across 12 rotations and five seeds, geometry reduces mean MSE from 0.2413 to
0.1128 AU$^2$ against a generic relational adapter, with four paired wins.
Their added parameter counts are 1,189,635 and 1,191,171. Five-seed nominal
MSE is 0.8596 for Plain NTP, 0.1366 with angular momentum (AM), and 2.0150
with a matched nonconserved radial target. AM wins five pairs against Plain
NTP and three against the radial control. With velocities and masses supplied
equally to both predictors, nominal MSE is $0.2301\pm0.3155$ without AM and
$0.1377\pm0.1515$ with AM (mean $\pm$ seed SD, three paired wins).
This nominal gain does not extend uniformly to shifts or conservation
(Appendix~\ref{app:revision_planetary}).
\end{revision}

\begin{revision}
\paragraph{TDV depends on the representation.}
TDV improves relative NTP's position accuracy in each original seed with encoding
fixed, and three of five seeds after extension. At the same weight, it increases pairwise NTP's position error while
reducing its angular-momentum drift (Table~\ref{tab:planetary_main}).
Better conservation and better forecasting therefore need to be checked
separately when combining temporal and geometric guidance.
Validation-selected TDV weights also fail to improve nominal mean MSE in four
independent confirmations (Appendix~\ref{app:revision_tdv}).
\end{revision}

\paragraph{Autonomous token forecasts under shifts.}
Relative + TDV reduces relative NTP's position MSE from \(1.6329\) to
\(0.7586\) AU\(^2\) at eccentricity 0.75 and from \(16.6076\) to \(4.9973\)
on the separate 400-frame test set.
Pairwise geometry is strongest on nominal forecasts, while relative + TDV
leads these controlled changes, using pretrained models without adaptation.

\subsection{Force learning and physical parameter transfer}
\label{sec:generalization}

\begin{figure}[!htb]
\centering
\includegraphics[width=\linewidth]{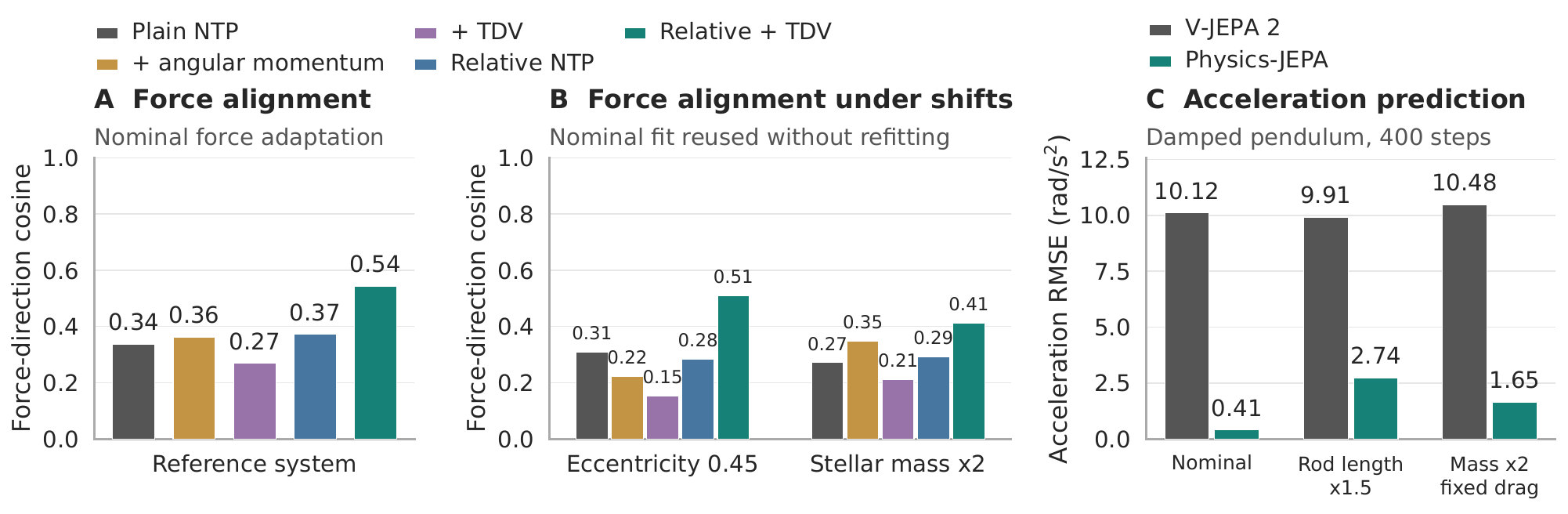}
\caption{Force-direction cosine after nominal adaptation (a) and parameter
changes without refitting (b), higher-is-better. Means cover eight planets
and two cadences, with 424 held-out labels per orbit, seed 1337.
(c) Damped-pendulum acceleration RMSE over 400 autonomous steps, 25 initial
conditions and three training seeds, lower-is-better.}
\label{fig:generalization}
\end{figure}

We next test force learning from limited labels, adapting each backbone with
the same budget. These tests use observed orbital positions.
Correct force arrows should turn with the planet and point toward its star.
Figure~\ref{fig:matched_force} compares equal-budget fitting within each
condition. Figure~\ref{fig:generalization}(a,b) tests transfer of the nominal
fit under parameter changes.

\input{paper_sections/generated/matched_force_main.tex}

\paragraph{Nominal force learning and parameter transfer.}
Across all 16 orbit and cadence combinations, relative + TDV improves
direction cosine over plain NTP from \(0.337\) to \(0.544\). Its directional
advantage persists on changed trajectories without refitting: \(0.308\) to
\(0.508\) at eccentricity 0.45 and \(0.271\) to \(0.413\) with doubled stellar
mass (Figure~\ref{fig:generalization}a,b). It has the highest mean among
all nine configurations, beating plain NTP in 13, 12, and 12 of the 16 cases.

\begin{revision}
\paragraph{Visual acceleration under parameter shifts.}
For the damped pendulum, Physics-JEPA reduces acceleration RMSE from
\(10.12\) to \(0.41\) rad/s\(^2\) at nominal parameters, from \(9.91\) to
\(2.74\) with a longer rod, and from \(10.48\) to \(1.65\) with doubled mass
at fixed drag (Figure~\ref{fig:generalization}c). These forecasts use the
trained head directly, without the changed parameters as inputs.
\end{revision}

\begin{revision}
\paragraph{New shift and inverse tests.}
Independent-body addition disrupts joint prediction. Factorized inference
preserves existing trajectories under the independent central-force assumption.
Fitting $g/L$ from a 32-frame decoded window lowers five-seed energy-head RMSE
from 0.7475 to 0.6223 rad, similarly to free and classical dynamics
(protocols in Appendices~\ref{app:revision_transfer}--\ref{app:revision_recovery}).
\end{revision}

%% file: paper_sections/generated/visual_main.tex
\begin{table}[!htb]
\caption{V-JEPA 2 baseline versus Physics-JEPA, separately trained with matched seeds. Forecasts: 400 steps on 25 initial conditions. State RMSE uses radians (spring: meters). Shared cells give total-system energy RMSE in joules for the double pendulum's upper and lower links. Single pendulums report mean $\pm$ sample SD over three seeds, and $\dagger$ marks one seed.}
\label{tab:visual_main}
\centering
\setlength{\tabcolsep}{1.5pt}
\begin{tabular}{@{}lrrrr@{}}
\toprule
System & Baseline RMSE & Physics RMSE & Baseline energy & Physics energy \\
\midrule
Simple pendulum & $0.8814 \pm 0.1892$ & $0.2212 \pm 0.0492$ & $3.6523 \pm 0.5831$ & $0.4106 \pm 0.0760$ \\
Damped pendulum & $1.1261 \pm 0.0174$ & $0.0446 \pm 0.0005$ & $5.0440 \pm 0.2389$ & $0.0987 \pm 0.0004$ \\
Spring--mass & ${3.1486}^{\dagger}$ & ${0.6905}^{\dagger}$ & ${5.7768}^{\dagger}$ & ${3.9664}^{\dagger}$ \\
Double pendulum, upper \(\theta_1\) & ${0.5709}^{\dagger}$ & ${0.1882}^{\dagger}$ & \multirow{2}{*}{${5.1660}^{\dagger}$} & \multirow{2}{*}{${0.5058}^{\dagger}$} \\
Double pendulum, lower \(\theta_2\) & ${1.1174}^{\dagger}$ & ${0.2600}^{\dagger}$ & & \\
\bottomrule
\end{tabular}
\end{table}

%% file: paper_sections/generated/planetary_main.tex
\begin{table}[!htb]
\caption{Autonomous planetary forecasts without adaptation. Mean $\pm$ sample SD across three seeds on 16 systems, with 32 observed and 128 future frames. Both metrics are lower-is-better. Drift uses a fixed physical scale. Relative and pairwise models share the coordinate encoding.}
\label{tab:planetary_main}
\centering
\setlength{\tabcolsep}{4pt}
\begin{tabular}{lrr}
\toprule
Model & Position MSE (AU$^2$) & Angular-momentum drift \\
\midrule
Plain NTP & $1.0755 \pm 1.6393$ & $0.9196 \pm 0.6293$ \\
Plain + angular momentum & $0.1437 \pm 0.0741$ & $0.4662 \pm 0.0929$ \\
Relative NTP & $0.2034 \pm 0.0864$ & $0.4537 \pm 0.0178$ \\
Relative + TDV & $0.1395 \pm 0.0542$ & $0.7274 \pm 0.0898$ \\
Pairwise NTP & $0.0630 \pm 0.0292$ & $0.6078 \pm 0.1268$ \\
Pairwise + TDV & $0.1439 \pm 0.0444$ & $0.5279 \pm 0.1198$ \\
\bottomrule
\end{tabular}
\end{table}

%% file: paper_sections/generated/matched_force_main.tex
\paragraph{Temporal learning improves force adaptation in relative coordinates.}
Among the displayed models in Figure~\ref{fig:matched_force}, relative + TDV
has the highest direction agreement and lowest vector error in all three
Earth conditions. Adding TDV to relative NTP increases direction cosine from
0.56 to 0.86, from 0.26 to 0.78, and from 0.40 to 0.79, respectively.
The relative-coordinate control isolates this temporal benefit.

\begin{revision}
\begin{figure}[!htb]
\color{revisionblue}
\centering
\includegraphics[width=\linewidth]{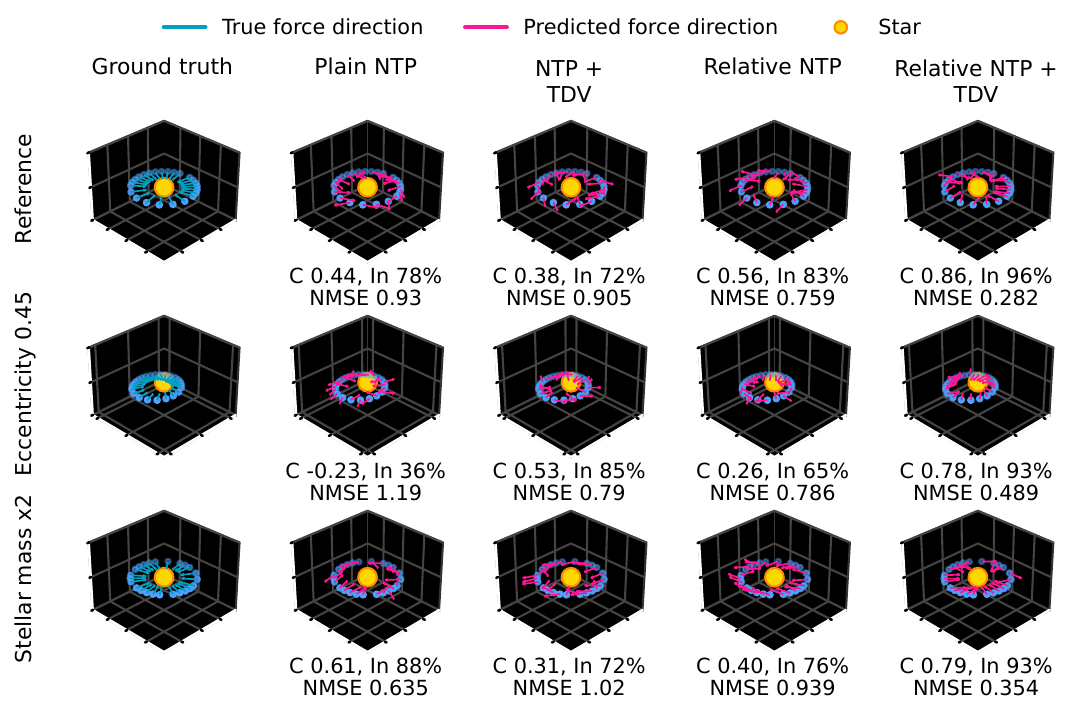}
\caption{Force directions after equal-budget adaptation on Earth trajectories,
using the orbital rendering of Figure~1 in \citet{vafa2025inductivebias}.
Each row is fitted separately from its pretrained checkpoint.
Arrows show unit directions at 24 matched positions, while metrics use
all 424 held-out labels. C is direction cosine and In is inward fraction,
both higher-is-better. Physical vector NMSE is lower-is-better.
Models use seed 1337. Earth was selected post hoc as an illustration.
Appendix~\ref{app:matched_force_adaptation} gives all 16 cases and
the full supervision comparison and selection protocol.}
\label{fig:matched_force}
\end{figure}
\end{revision}

%% file: paper_sections/09_discussion.tex
\section{Discussion and Limitations}
\label{sec:discussion}

\paragraph{Match the structure to the prediction task.}
Visual mechanics generates the next state directly.
Planetary supervision and geometry guide the token predictor that generates
coordinates. Both require checking trajectories alongside physical scores.
Energy can stay constant at the wrong amplitude, and a phase portrait can
trace the right curve at the wrong speed. Joint trajectory and physical
metrics reveal both errors.

\begin{revision}
\paragraph{Scope and next steps.}
Ordered attention and quantization limit end-to-end equivariance.
Matched controls distinguish transition-level visual benefits and geometric
robustness from added supervision and capacity (Appendix~\ref{app:revision_controls}).
Body-count shifts and parameters ambiguous from two images remain difficult.
New visual controls use five seeds. Original Relative+TDV force results use
one pretraining seed, while five-seed Plain/AM adaptation is mixed.
Force direction is mass-independent at fixed position. Further tests must
assess mass-scaling of force magnitude and integration error. Training and
data efficiency require matched learning curves.
\end{revision}

%% file: paper_sections/10_conclusion.tex
\section{Conclusion}

\begin{revision}
When observations stop, a predictive model must sustain motion using what it
has learned. Compared with the V-JEPA 2 baseline, learned mechanics improves
trajectories, acceleration and energy accuracy across four visual systems.
With labels and RK4 matched, energy heads improve simple- and double-pendulum
motion, while spring and damped-pendulum gains mainly concern consistency.
Angular-momentum supervision improves forecasting and momentum stability over
position-input Plain NTP. Equal velocity/mass inputs retain a nominal
forecasting benefit, but not improved conservation.
Pairwise geometry improves Relative NTP and rotation robustness over a
capacity-matched generic adapter. Temporal benefits depend on the task
and representation.
\end{revision}

\begin{revision}
Phase-space plots and joint trajectory/physical metrics show these effects.
Within-family controls distinguish structure from added capacity and supervision.
A learned mechanical head improves visual
rollout by advancing the decoded state. An auxiliary objective or geometric
adapter improves planetary rollout by changing the token predictor itself.
Parameter and horizon tests locate where gains persist. The practical lesson
is to connect physical guidance to the computation producing future states,
then evaluate the motion it generates.
\end{revision}

%% file: paper_sections/12_statements.tex
\subsection*{AI use statement}
Generative AI tools assisted with experimental design, implementation and
debugging, analysis of results, literature search, figure preparation, and
manuscript drafting and editing. Verification includes automated tests,
checks of evaluation artifacts and source records, and visual inspection of
figures. The authors are responsible for the final methods, interpretation,
citations, and presentation.

\subsection*{Reproducibility statement}
Section~\ref{sec:experimental_protocol} specifies the data, training budgets,
evaluation horizons, and aggregation rules. The appendix provides objective
definitions, physical metrics, complete comparison tables, and additional
implementation details. Figures and tables are generated from saved
predictions with recorded configurations and source hashes.

%% file: paper_sections/11_appendix.tex
\section{Mathematical and Implementation Details}
\label{sec:appendix}

The appendix specifies the training objectives, physical metrics, and
evaluation scenarios, followed by the complete model comparisons and
seed-level measurements supporting the main text.
Tables and figures are generated from saved numerical forecasts, with
recorded configurations, checkpoint hashes, and checks for missing or nonfinite
values. Aggregates include all evaluation systems and available training
seeds for each comparison.

\begin{revision}
\paragraph{Appendix guide.}
Appendices~\ref{sec:appendix}--\ref{app:generalization} give the original
methods, full model comparisons and generalization matrices.
Appendix~\ref{app:revision_controls} adds the matched controls and diagnostics:
visual dynamics in~\ref{app:revision_visual}, planetary capacity and
equal-information comparisons in~\ref{app:revision_planetary}, temporal-objective
selection in~\ref{app:revision_tdv}, body-count interventions
in~\ref{app:revision_transfer}, coefficient recovery
in~\ref{app:revision_recovery}, observation correction
in~\ref{app:revision_correction}, the image-based baseline
in~\ref{app:revision_named}, and force adaptation in~\ref{app:revision_force}.
Aggregate results precede their individual-seed tables.
\end{revision}

\subsection{Gradient flow and visual consistency}
\label{app:background_equations}

The main text defines the predictive objectives. Here we detail how the
visual models generate forecasts and how their losses update each component.
Let \(h_t\) be encoded image features at frame \(t\), and let \(P\) denote
the autoregressive use of predictor \(g_\phi\). Decoder \(D\) maps two
successive feature vectors to physical state \(y=(q,v)\).
The operator \(\Phi^\psi_{\Delta t}\) is one fourth-order Runge--Kutta
(RK4) step of the learned mechanical equations with parameters \(\psi\)
over frame interval \(\Delta t\). Hats mark predictions and \(k\) counts
future steps:
\begin{equation*}
\begin{aligned}
 \text{JEPA:}\quad &\hat h_{t+1}=P(\hat h_{t-1},\hat h_t),
 &\hat y_{t+1}=D(\hat h_t,\hat h_{t+1}),\\
 \text{Mechanics:}\quad &\hat y_t=D(h_{t-1},h_t),
 &\hat y_{t+k}=\Phi^\psi_{\Delta t}(\hat y_{t+k-1}).
\end{aligned}
\end{equation*}
The first predicted features use the observed image pair as their history.
Each later step uses previous predictions.

\begin{revision}
Baseline V-JEPA 2 (condition A)
trains visual prediction and state decoding. Physics-JEPA (condition B)
also trains a mechanical head. Its mechanical losses leave the visual
branch unchanged, with separate gradient clipping preserving the visual
updates. The consistency-coupled model (condition C) adds a
stopped-gradient one-step mechanical target to visual
prediction:
\begin{equation*}
 \mathcal L_{\mathrm{cons}}
 =\left\|\hat y^{\mathrm{latent}}_{t+1}
 -\operatorname{sg}\!\left[\Phi^\psi_{\Delta t}
             (\operatorname{sg}[\hat y_t])\right]\right\|_2^2.
\end{equation*}
Here \(\hat y^{\mathrm{latent}}_{t+1}\) is the state decoded from predicted
features. This argument carries gradients to the visual predictor.
The stopped-gradient target leaves the
mechanics head's supervised update unchanged.
\end{revision}

\subsection{Conserved quantities and force supervision}
\label{app:physical_targets}

\begin{revision}
For relative displacement \(r\), velocity \(v\), planet mass \(m\), and
stellar mass \(M\), define reduced mass \(\mu=mM/(m+M)\).
The finite-mass two-body training targets are
\begin{align*}
 H&=\tfrac12\mu\|v\|^2-GmM/\|r\|,\\
 L_z&=\mu(r_xv_y-r_yv_x),\\
 F&=-GmM r/\|r\|^3.
\end{align*}
The fixed-star limit gives \(\mu\simeq m\) when \(m\ll M\).
Targets use aligned state records. A masked regression loss
averages only available physical labels. Force-law supervision calculates the
force target from physical state, while force auxiliary supervision reads
the corresponding numerical target. These objectives supervise physical
values during training. At generation, NTP uses its learned token
distribution without an analytic conservation constraint.
\end{revision}

\subsection{Equivariance of the geometric branch}
\label{app:equivariance}

\begin{revision}
Under \(r_i'=Qr_i+a\), \(Q\) is a rotation or reflection
(\(Q^\mathsf TQ=I\)) and \(a\) a translation. Relative vectors satisfy
\(r_{ij}'=Qr_{ij}\). Relative motion features transform in the same way.
Their squared norms and inner products are invariant, so
\(u_{ij}'=u_{ij}\). The scalar coefficient networks therefore return the
same values and
\begin{equation*}
 d_i'=\sum_{j\ne i}\alpha_\psi(u_{ij})Qr_{ij}
       +\beta_\psi(u_{ij})Qw_{ij}
 =Qd_i.
\end{equation*}
Summation is invariant to reordering source messages. This proof concerns the
continuous proposal with matched source identities. The surrounding ordered
transformer, feature injection, quantization, and autoregressive decoding
are evaluated empirically as a complete system. The construction specifies
transformation behavior without specifying an inverse-square exponent or a
conserved energy function. The shared message MLP has one hidden layer of
width 1536 with a GELU activation, producing one embedding-width message and
two scalar coefficients per pair. For coordinate \(a\), proposal bias at
bin \(k\) is
\begin{equation*}
 B_{a,k}(d_i)=-\frac12\left(\frac{c_k-50d_{i,a}}{64\,\Delta c}\right)^2,
 \qquad \Delta c=100/7000\ \mathrm{AU},
\end{equation*}
where \(c_k\) is the signed displacement bin center. Special-token entries
receive the coordinate-bias floor, and the proposal mask disables the bias
when no complete frame is available. This bias and the hidden message share
the learned scalar gate in the following interface.
For row \(n\) predicting body \(i\), \(e_n\) is the original input embedding,
\(e'_n\) the updated embedding, and \(\ell_n\) the output logits:
\begin{equation*}
 e'_n=e_n+g\eta_i,\qquad
 \ell_n=W\,\operatorname{GPT}(e'_{\leq n})+g B(d_i).
\end{equation*}
Here \(\operatorname{GPT}\) maps the causal embedding history \(e'_{\leq n}\)
to a hidden state, \(W\) is the output projection, and \(g\) is a learned
scalar gate initialized to zero. The message \(\eta_i\) has the embedding
dimension. The vector \(B(d_i)\) contains the biases defined above for
the displacement proposal \(d_i\). Both connections learn through NTP.
\end{revision}

\subsection{Learned mechanics and supervised dynamics}
\label{app:mechanics}

\begin{revision}
Hamiltonian mechanics in canonical coordinates and Lagrangian mechanics in
velocity coordinates satisfy the equations below. Here \(q\) is position,
\(p\) canonical momentum, \(v\) velocity, \(H\) the Hamiltonian energy,
and \(L\) the Lagrangian:
\begin{align*}
 \dot q&=\partial_pH,& \dot p&=-\partial_qH,\\
 \frac{\mathrm d}{\mathrm dt}\partial_vL-\partial_qL&=0.
\end{align*}
The implementation uses velocity state \((q,v)\).
For one degree of freedom it uses unit-inertia kinetic energy and a learned
potential. The double-pendulum head uses
\begin{align*}
 T(q,v)={}&\tfrac12(m_1+m_2)l_1^2v_1^2
       +\tfrac12m_2l_2^2v_2^2\\
       &+m_2l_1l_2\cos(q_1-q_2)v_1v_2.
\end{align*}
Here \(q_1,q_2\) are upper- and lower-link angles, \(v_1,v_2\) their angular
velocities, \(m_1,m_2\) their masses, and \(l_1,l_2\) their lengths.
With learned potential \(V_\psi\) and dissipation function \(R_\psi\), the
Euler--Lagrange equation is
\begin{equation*}
 \frac{\mathrm d}{\mathrm dt}\partial_v[T(q,v)-V_\psi(q)]
 -\partial_q[T(q,v)-V_\psi(q)]=-\partial_v R_\psi(q,v).
\end{equation*}
Writing \(L=T-V_\psi\), automatic differentiation constructs inertia matrix
\(M=\partial_v^2L\) and mixed-derivative matrix \(C=\partial_q\partial_vL\),
then solves
\begin{equation*}
 M\dot v=\partial_qL-Cv-\partial_vR_\psi.
\end{equation*}
The coupled kinetic matrix and training-system parameters are supplied.
The potential remains learned. Damping uses positive quadratic Rayleigh
coefficients, ensuring nonpositive dissipative power. The heads receive
numerical acceleration supervision and a power-balance objective:
\begin{align*}
 \mathcal L_{\mathrm{dyn}}&=\|f_\psi(y)-\dot y\|^2,\\
 \mathcal L_{\mathrm{power}}&=
    |\nabla_y E_\psi(y)^\mathsf T\dot y-\dot E_\psi^{\mathrm{expected}}|^2.
\end{align*}
Here \(f_\psi(y)\) is the learned state derivative,
\(\dot y\) the simulated derivative, and \(\nabla_y E_\psi\) the energy
gradient with respect to state. The conservative expected power is zero,
and the damped expected power follows
the Rayleigh term. A separate Lagrangian head receives acceleration
supervision. Every reported assisted trajectory integrates the energy-based
head with RK4 from an image-derived initial state.
\end{revision}

\subsection{Training settings}
\label{app:training_contracts}

Visual training initial conditions are independent uniform draws. The simple
pendulum uses \(\theta_0\in[-1.2,1.2]\) rad and
\(\omega_0\in[-1.6,1.6]\) rad/s. The damped pendulum uses
\([-1.5,1.5]\) rad and \([-2,2]\) rad/s. Both have nominal length 1 m
and gravity 9.81 m/s\(^2\), with damped angular coefficient 0.3 s\(^{-1}\).
The spring uses mass 1 kg, stiffness 3 N/m, \(x_0\in[-2.5,2.5]\) m,
and \(v_0\in[-3,3]\) m/s. Each double-pendulum arm has length 1 m and
mass 1 kg, with independent initial angles in \([-1.2,1.2]\) rad and
angular velocities in \([-1.5,1.5]\) rad/s. Entire simulated episodes
remain within their split when extracting overlapping three-frame windows.

Visual JEPA loss has weight 1. Position and velocity losses each have weight
5, or 10 per coordinate for the double pendulum. The energy-based and
Lagrangian derivative losses each have weight 0.05 in B and C. Power loss
has weight 0.05 for conservative systems and 0.3 for the damped pendulum.
These terms ramp linearly from epochs 5 to 20, or 10 to 40 for the double
pendulum. C adds consistency with weight 2 under the same ramp.
The potential MLP has three width-256 tanh layers. The single-DOF head feeds
position and a zero velocity into this network, ensuring potential energy
depends on position alone. Context-encoder targets use EMA decay 0.999.
Training predicts one future frame per window, while evaluation recursively
generates the entire forecast without fresh observations.
Visual optimization uses AdamW with learning rate \(3\times10^{-4}\),
cosine decay, and batch size 32.

Planetary auxiliary regressions attach to the currently observed body row
and exclude star and metadata rows where physical labels are masked. The
main temporal runs use first differences of quantized star-relative
coordinates, omit star pairs, and retain the full causal attention context.
All combinations use the common auxiliary weight 0.1. Planetary validation selects the best
base-task checkpoint within the fixed training budget, using 10 validation batches
every 250 updates. The adapter adds exactly 1,189,635 parameters.
Visual evaluation
uses the final epoch, consistently across A, B, and C.

\subsection{Metrics and aggregation}
\label{app:force_metrics}

\begin{revision}
Planetary position MSE excludes the star and averages over the forecast
window, planet coordinates, and systems. For an angular prediction, visual
RMSE uses \(\operatorname{atan2}(\sin\delta,\cos\delta)\), with
\(\delta=\hat q-q\). Observed initialization frames are omitted. The main
visual energy-value error compares physical energy evaluated on the predicted
and numerical trajectories:
\begin{equation*}
 \operatorname{RMSE}_E=
 \sqrt{\frac{1}{ST}\sum_{s=1}^{S}\sum_{t=1}^{T}
  [E_{\mathrm{phys}}(\hat y_{s,t})-E_{\mathrm{phys}}(y_{s,t})]^2}.
\end{equation*}
Here \(S\) counts evaluation systems and \(T\) future steps for one trained
model. State RMSE follows the same pooling order, using circular angular
errors where appropriate. Training seeds are aggregated only afterward.
Double-pendulum energy is computed jointly from both arms.
Drift instead measures changes relative to the
predicted cutoff energy, including accumulated dissipation for damped motion.
Both are needed because a conserved but incorrect orbit can have low drift.
\end{revision}

For planetary angular momentum, the evaluator forms relative velocities by
finite differences and computes \(\hat L_t=\mu(\hat r_{x,t}\hat v_{y,t}
-\hat r_{y,t}\hat v_{x,t})\). The reported drift for one system is
\begin{equation}
 D_L=\frac{1}{T}\sum_t\frac{|\hat L_t-\hat L_0|}{S_L},\qquad
 S_L=\max\!\left(|L^{\mathrm{ref}}_0|,
 \mu\sqrt{G(m+M)\|r^{\mathrm{ref}}_0\|},\epsilon\right).
 \label{eq:angular_momentum_drift}
\end{equation}
The fixed scale uses the true initial context state.
Here \(L^{\mathrm{ref}}_0\) and \(r^{\mathrm{ref}}_0\) are
the true initial angular momentum and star-relative position, \(m,M\) the
planet and stellar masses, \(\mu=mM/(m+M)\) their reduced mass, and
\(\epsilon\) a positive numerical floor. Centered differences
estimate interior velocities, with second-order one-sided differences at
the endpoints. Evaluation averages system drifts within each trained model
before aggregating seeds. Low drift measures stability around the predicted
initial value, so it is interpreted alongside trajectory error.

\begin{revision}
For star-to-planet vector \(r_i\), true force \(F_i\), and estimate
\(\hat F_i\), the supplementary force geometry diagnostics include
\begin{align*}
 C_{\mathrm{dir}}&=\frac1n\sum_i
   \frac{\hat F_i^\mathsf TF_i}{\|\hat F_i\|\|F_i\|},\\
 P_{\mathrm{in}}&=\frac1n\sum_i\mathbb 1[\hat F_i^\mathsf Tr_i<0].
\end{align*}
Direction and inward fraction are higher-is-better. Vector error is
lower-is-better. Zero-vector and nonfinite
cases are counted explicitly. Normalized acceleration error uses a
zero-acceleration reference. Forecasting tables report variation across
training seeds. The force-adaptation table reports variation across orbit
and cadence cases for one pretrained seed. Captions identify the aggregation unit.
\end{revision}

\section{Additional Visual Results}
\label{app:visual}

\subsection{Physical metrics and consistency coupling}
The nominal results use the same numerical truth across matched conditions.
The main table compares the V-JEPA 2 baseline (A) with Physics-JEPA (B).
Phase portraits use initial-condition grid indices 6 and 18 at seed 42,
chosen before examining model performance.
The table below also compares learned mechanics with latent autoregression
within Physics-JEPA, from the same initial images.
The consistency-coupled model (C) tests additional dynamics-to-visual consistency.
It improves latent angle forecasting in one of three seeds for each
single-pendulum system. Its structured forecasts remain close to B's.
\input{paper_sections/generated/visual_physics}
\input{paper_sections/generated/visual_coupling}

\begin{figure}[!htbp]
\centering
\includegraphics[width=.9\linewidth]{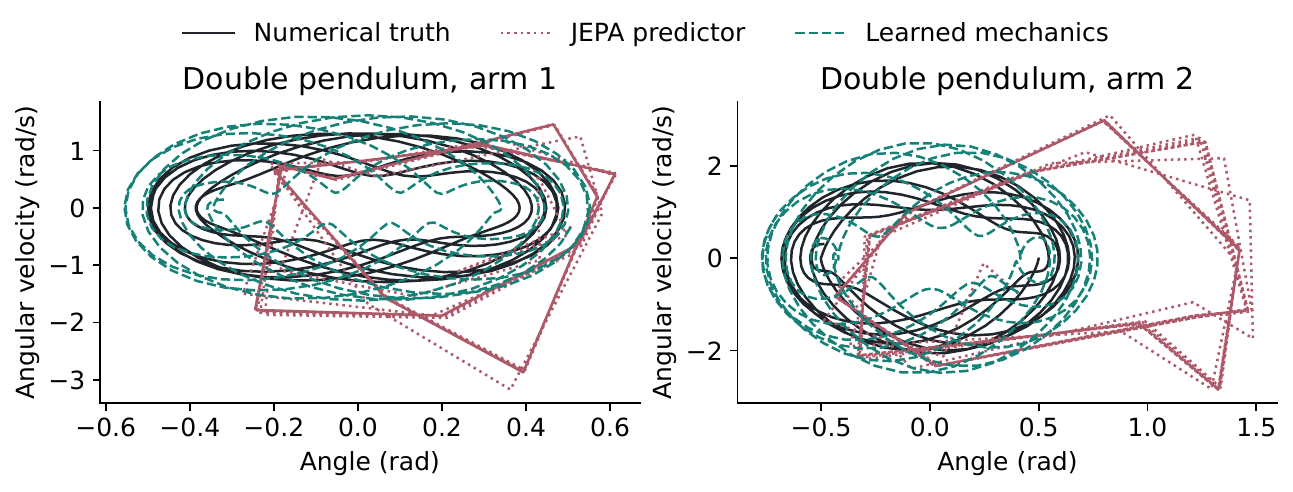}
\caption{Both double-pendulum coordinates for held-out grid index 18,
condition B, seed 42. The trajectories use the same initial image pair and
400 future targets. The nominal table aggregates all 25 initial conditions.}
\label{fig:double_phase_space}
\end{figure}

\subsection{Matched local and autonomous prediction}
This test supplies true image pairs before each fresh-frame prediction,
while autonomous prediction receives only the initial pair. Both paths score
the same 160 targets in 25 systems. Mean fresh-frame versus autonomous latent
RMSE is \(0.0968/0.8812\) for simple pendulum,
\(0.0247/1.1571\) for damped pendulum, \(0.2044/3.0893\) for spring,
and \(0.0339/0.5657\) for double-pendulum arm one. Fresh observations correct
state-estimation as well as transition error, so this test measures their
combined effect.

\section{Additional Planetary Results}
\label{app:planetary}

\subsection{Autonomous trajectories and motion vectors}
\begin{figure}[!htbp]
\centering
\includegraphics[width=\linewidth]{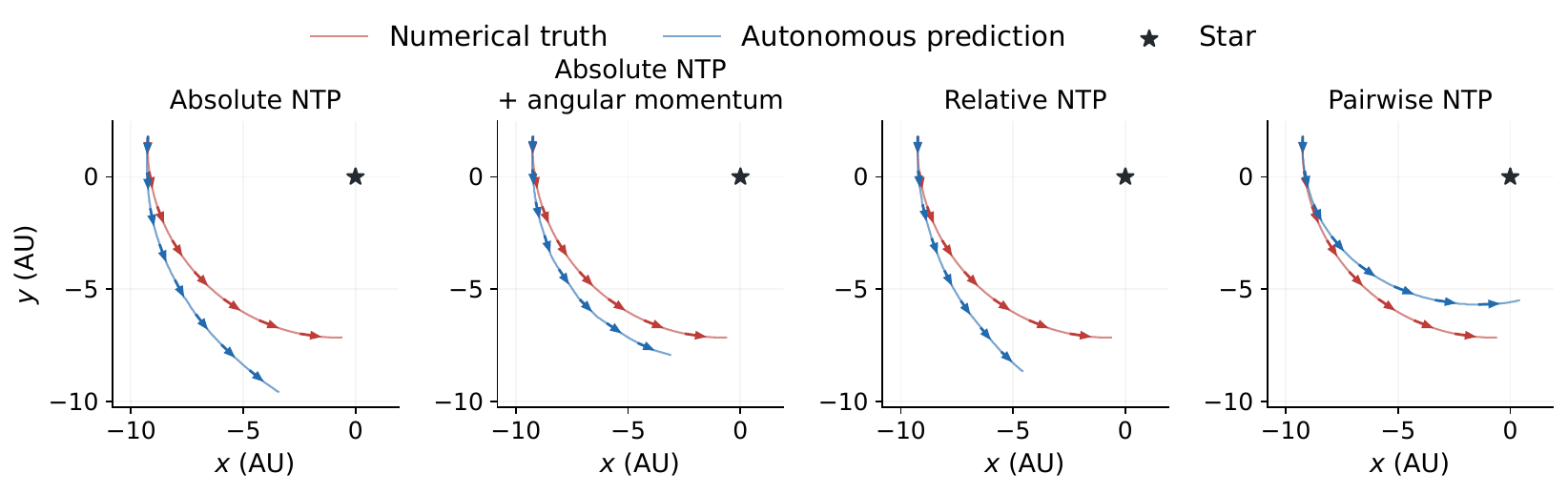}
\caption{Predicted planetary motion after 32 observed frames. Red and blue
curves show numerical truth and autonomous forecasts over 128 future frames.
Arrows show eight-frame displacement in AU at every sixteenth frame, anchored
on each trajectory with identical scales across panels. All panels use
held-out system 0 and training seed 1338, without post-training adaptation.
Table~\ref{tab:planetary_main} and Table~\ref{tab:planetary_motion_vectors}
aggregate all 16 systems and three seeds.}
\label{fig:planetary_motion_vectors}
\end{figure}

\begin{figure}[!htbp]
\centering
\includegraphics[width=\linewidth]{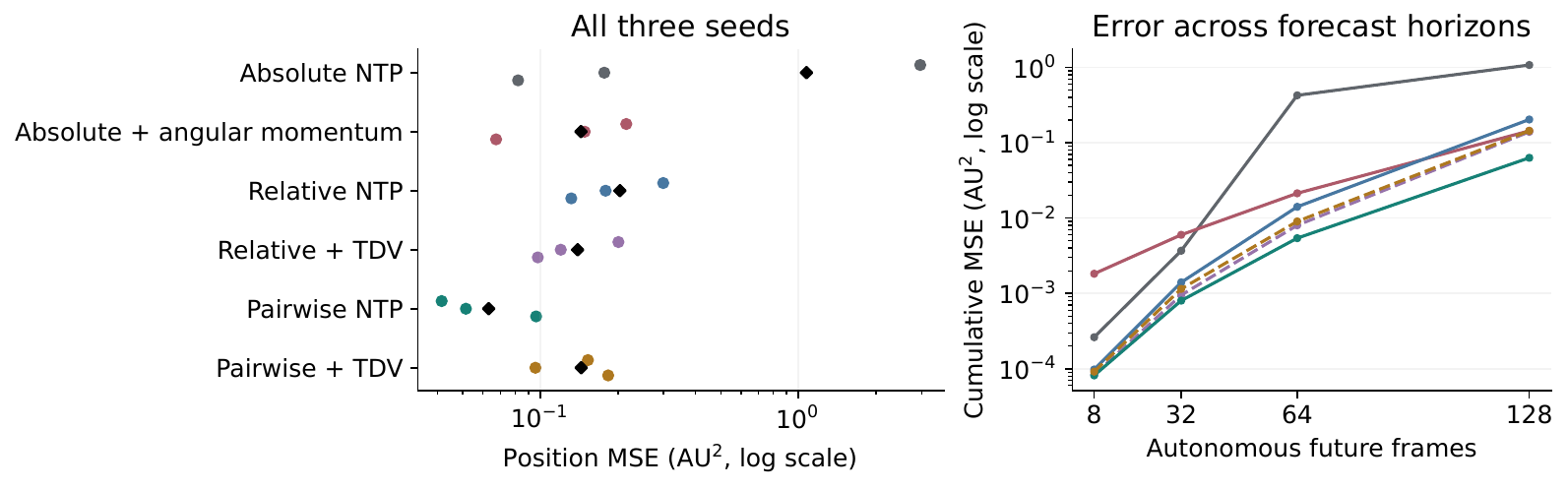}
\caption{Autonomous planetary forecasting. Left: all three training seeds,
with black diamonds indicating means. Right: cumulative error at the four
evaluated horizons, averaged over seeds. Colors match across panels.
The error axes are logarithmic. Angular-momentum supervision and pairwise
geometry improve their respective matched baselines in every seed.}
\label{fig:planetary_forecasting}
\end{figure}

Figure~\ref{fig:planetary_forecasting} shows all training seeds and the
growth of cumulative position error. Figure~\ref{fig:planetary_motion_vectors}
uses the first held-out system and the middle listed training
seed. No model-dependent choice of system, seed, or time window is made.
The entire autonomous forecast is displayed, with unnormalized arrows
\(d_t=x_{t+8}-x_t\) at every sixteenth frame. No smoothing is applied.

\begin{revision}
Table~\ref{tab:planetary_motion_vectors} evaluates all 120 overlapping
eight-frame displacements in each of the 16 systems. For each seed,
\begin{equation*}
 E_{\mathrm{motion}}=
 \frac{\sum_{s,b,t}\|\hat d_{s,b,t}-d_{s,b,t}\|^2}
      {\sum_{s,b,t}\|d_{s,b,t}\|^2},
 \qquad d_{s,b,t}=x_{s,b,t+8}-x_{s,b,t}.
\end{equation*}
The star is excluded, and only future frames enter the calculation.
The fixed eight-frame interval is the shortest horizon reported in the
existing forecast evaluation. Cadence is identical across models for each
system, though it differs between systems. These displacement vectors are
measured in AU over eight frames. Direction cosine is computed
only for nonzero predicted and reference vectors, with coverage reported.
Normalized vector error includes zero predictions and measures magnitude
as well as direction. This diagnostic describes finite-interval motion
and is distinct from the acceleration and frozen-readout tests below.
\end{revision}

\input{paper_sections/generated/planetary_motion_vectors}

\subsection{Complete objective comparison}
Table~\ref{tab:planetary_full} includes every current objective condition.
The primary contrasts have three seeds. Additional force, energy, and
pairwise-invariant combinations have one seed and supply exploratory ablations.
Plain NTP conditions with TDV use the original auxiliary coordinate encoding.
The relative and pairwise comparisons hold the revised encoding fixed.
\input{paper_sections/generated/planetary_full}

\subsection{Conservation and frozen readouts}
Table~\ref{tab:angular_momentum} gives the matched-seed conservation evidence
for angular-momentum supervision. All three seeds improve relative to
absolute NTP, matching the direction of the position-error improvement.
\input{paper_sections/generated/angular_momentum}

Pairwise NTP's mean normalized energy and angular-momentum drift are
\(0.7696\) and \(0.6078\). Adding TDV lowers these to \(0.3880\) and
\(0.5279\), while increasing position MSE. Frozen linear acceleration
readouts are trained on disjoint systems without backbone updates and
evaluated on 100 held-out systems with 25 positions per system.
Their normalized MSEs for relative NTP, relative + TDV, pairwise NTP, and
pairwise + TDV are respectively \(0.9866\), \(0.9935\), \(0.9876\), and
\(0.9942\). Direction cosines are \(0.2194\), \(0.2124\), \(0.1919\), and
\(0.2004\). These results distinguish forecasting improvement from
recovery of a reusable acceleration field.

\subsection{Pretrained force-head control}
\label{app:native_force}

Force-supervised models also predict forces directly from their pretrained
heads, without further fitting. Each head receives the observed planet
coordinates and causal history. A control on an Earth-mass orbit with
eccentricity 0.45, a one-solar-mass star, and cadence 0.0189 years illustrates
why direction scores require care. This case was selected after inspecting
48 planet, cadence, and parameter combinations for consistent improvements
from TDV across three seeds. It contains 499 evaluated positions.

For the force-supervised TDV models, replacing predictions by each model's
mean force on this test orbit gives direction cosine 0.3882, compared with
0.3729 for the original predictions. Magnitude calibration is also poor.
Thus a fixed orientation can explain the directional score on this selected
orbit. These pretrained heads are evaluated separately from the fresh heads
and full-backbone adaptation used in the main force comparison.

\subsection{Matched sparse force adaptation}
\label{app:matched_force_adaptation}

Figure~\ref{fig:matched_force} compares four pretrained configurations after
the same force-adaptation procedure. Each checkpoint receives a fresh linear
force head, with the retained backbone trainable. Existing auxiliary heads
are removed and their losses disabled. The pairwise variants retain their
embedding branch, while the coordinate-logit proposal is inactive for force
regression. AdamW uses learning rate \(6\times10^{-4}\), batch size 8,
gradient clipping 1, and 1,000 updates. All configurations use pretrained
seed 1337 and the same head initialization and label splits.

\begin{revision}
Each fit contains eight named-planet two-body orbits at both 0.0189 and
0.486 years per frame. Of 499 available labels per orbit, 25 fit the model,
50 select the checkpoint, and 424 provide held-out evaluation. One scalar
force normalization is computed from fitting labels only and applied to both
coordinates. Metrics use physical vectors. For each orbit, vector NMSE is
\begin{equation*}
 E_F=\frac{\sum_{t\in\mathcal H}\|\hat F_t-F_t\|^2}
               {\sum_{t\in\mathcal H}\|F_t\|^2},
\end{equation*}
where \(\mathcal H\) is the held-out label set. The direction cosine and
inward fraction follow Appendix~\ref{app:force_metrics}. The arrows alone
show orientation, while vector NMSE also evaluates magnitude.
\end{revision}

There are three fitting conditions: reference parameters, eccentricity 0.45,
and doubled stellar mass. Each starts from the original pretrained checkpoint.
The reference and doubled stars have one and two solar masses, outside
the pretraining mass range. These tests measure force adaptation on observed
orbits. Autonomous forecasting and fixed-model parameter transfer are
separate evaluations. The two relative variants share the corrected
integer-coordinate encoding. Absolute TDV retains its original auxiliary encoding,
so comparisons across representations also include that implementation choice.

The main figure uses Earth at the finer cadence and shows all three
conditions with relative-coordinate and temporal controls.
It was chosen after examining the results. Twenty-four displayed positions
are selected for spatial coverage from the shared held-out set, starting at
its first point. Metrics always include all 424 held-out positions.
Table~\ref{tab:matched_force_all_models} includes every pretraining variant
on these same Earth cases. Table~\ref{tab:matched_force_catalog} checks the
direction of improvement across all 48 planet, cadence, and condition cases.
These related cases supply descriptive comparisons from one training seed.
\input{paper_sections/generated/matched_force_all_models}
\input{paper_sections/generated/matched_force_catalog}

\subsection{Aggregate force directions and parameter transfer}
\label{app:force_alignment_transfer}

Figure~\ref{fig:generalization}(a,b) uses the model fitted on reference
parameters for every condition. This differs from the separately fitted rows
of Figure~\ref{fig:matched_force}. Each mean includes all eight planet
identities and both cadences with equal weight. The 424 held-out labels per
orbit are unchanged across models. The shifted conditions use no additional
fitting or checkpoint selection. Table~\ref{tab:force_alignment_transfer}
reports all nine variants. The variation shown is across the 16 related
orbit/cadence cases, conditional on pretraining seed 1337.
\input{paper_sections/generated/force_alignment_transfer}

Relative + TDV improves direction cosine over plain NTP on 13 of 16 reference
cases, 12 of 16 eccentricity cases, and 12 of 16 stellar-mass cases. The
corresponding median paired improvements are 0.2464, 0.1417, and 0.1921.
These tests support improved force directions after adaptation. Force
magnitude rankings depend on the condition and aggregation. For example,
pooled reference vector NMSE is 0.4465 for plain NTP and 0.6638 for relative
+ TDV, whereas their means of per-orbit NMSE are 112.25 and 19.55.
The pooled score gives more weight to orbits with larger true forces.

\subsection{Uncertainty across held-out systems}
The main tables report training-seed variation. Table~\ref{tab:planetary_system_intervals}
separately measures variation across the 16 nominal evaluation systems.
For each system, we average the paired treatment-minus-baseline position MSE
across the three fixed training seeds. We then resample those 16 system-level
differences with replacement 10,000 times and report percentile 95\% bootstrap
intervals for their mean. Negative differences favor the treatment. This
preserves the pairing and the within-trajectory dependence across frames.
The intervals are conditional on the trained checkpoints and do not replace
replication over additional training seeds.
\input{paper_sections/generated/planetary_system_intervals}

\subsection{Coordinate encoding and sampling}
Relative and pairwise models use integer-bin displacements and accumulation
throughout training and generation.
Quantization alone produces position MSE of approximately \(0.000034\)
AU\(^2\) on the original test set, below current model errors.

The training generator defines a parametric population by sampling orbital
parameters independently for each system. Each training system contains one
planet. All bodies in a system share a regular physical-time grid.
The cadence varies across systems, so equal token counts need not cover equal
fractions of an orbit. Instantaneous distance occupancy also differs from
the semi-major-axis distribution, since eccentric orbits spend unequal time
at different radii. Distance-error comparisons therefore require stratifying
by cadence and forecast horizon. Controlled eccentricity interventions are
reported separately from broader distribution changes.

\section{Complete Generalization Matrices}
\label{app:generalization}

\subsection{Seed-level controlled comparisons}
\label{app:generalization_seeds}
\begin{figure}[!htbp]
\centering
\includegraphics[width=\linewidth]{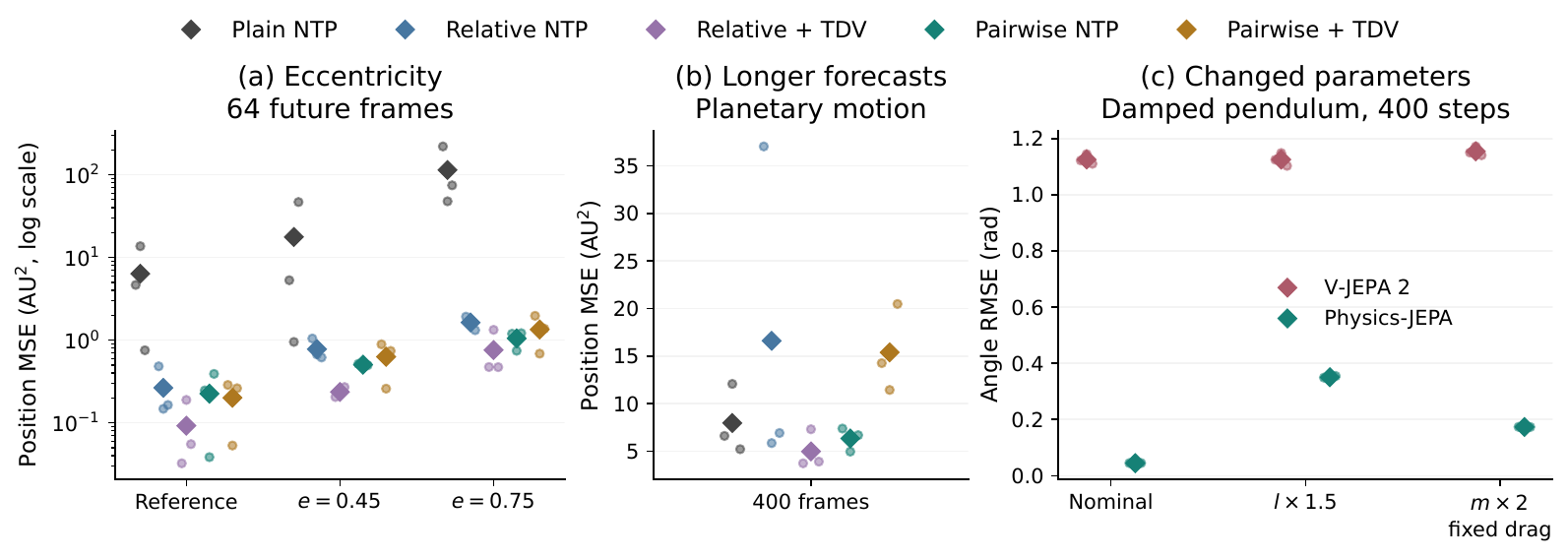}
\caption{Seed-level forecasting errors. Circles show the three individual
training seeds and diamonds show their arithmetic mean. Planetary panels
measure position MSE, and the visual panel measures angle RMSE.
Panel (a) uses a log scale. The 400-frame planetary test set is separate from
the 64-frame eccentricity sets. Each visual score covers 400 future
steps on 25 initial conditions.}
\label{fig:generalization_seeds}
\end{figure}

Figure~\ref{fig:generalization_seeds} shows seed-level forecasting errors
for the controlled eccentricity and horizon tests, together with visual
position errors under parameter changes. These supplement the force-direction
and visual acceleration metrics in Figure~\ref{fig:generalization}.
The visual comparison uses independently trained baseline and Physics-JEPA
models with matched initial images and numerical truth.

\subsection{Planetary interventions}
The paired suite reuses random seeds and nuisance draws while intervening on
mass, eccentricity, horizon, or body composition. Physical states and system
identifiers can change under these interventions. Its reference test set differs
from the original-distribution test set in the main table. Matched
forecasts normally span 64 frames, with 400 for the long-horizon condition.
The standard suite includes broader, compound shifts and coordinate changes.
Both suites are reported below without pooling their means.

Relative + TDV performs well on several controlled shifts. Larger body-count
changes remain difficult. The matched four-planet case also changes mass
distributions, and persistence scores \(97.1951\) AU\(^2\), below the mean
error of all four displayed models. These tests compose independent Kepler
orbits without interplanetary forces.
\input{paper_sections/generated/planetary_matched}
\input{paper_sections/generated/planetary_standard}

\subsection{Visual interventions}
Each visual table reports every condition-B scenario. All forecasts span
400 future steps and use the numerical truth associated with that scenario.
Mass, gravity, and length values are not explicitly passed to the learned
dynamics at evaluation. A visible-length shift changes the rendering as well
as the dynamics, while a hidden-length shift preserves nominal visual length.
Mass-null and matched scaling controls test evaluator consistency. They are
not counted as independent demonstrations of new physical behavior.

The structured head retains useful advantages in several damped and spring
conditions. Higher gravity for simple pendulum and a longer first rod for
double pendulum illustrate cases where its nominal advantage does not persist.
These tests show how particular physical changes affect the learned
transition.
\input{paper_sections/generated/visual_shifts_simple_pendulum}
\input{paper_sections/generated/visual_shifts_real_pendulum}
\input{paper_sections/generated/visual_shifts_spring_block}
\input{paper_sections/generated/visual_shifts_double_pendulum}

%% file: paper_sections/generated/visual_physics.tex
\begin{table}[!htbp]
\caption{Acceleration RMSE and energy drift/balance RMSE for the same 400-step forecasts. Acceleration is in rad/s$^2$ (spring in m/s$^2$). Energy is in joules. The damped system includes dissipated energy. Double-pendulum acceleration pools both arms, and its energy drift is a whole-system metric. $\dagger$ indicates one seed.}
\label{tab:visual_physics}
\centering
\setlength{\tabcolsep}{2pt}
\begin{tabular}{@{}lrrrr@{}}
\toprule
System & Latent accel. & Structured accel. & Latent drift & Structured drift \\
\midrule
Simple pendulum & $33.9603 \pm 24.5465$ & $1.8284 \pm 0.3823$ & $3.4920 \pm 0.5602$ & $0.0002 \pm 1.61\times10^{-5}$ \\
Damped pendulum & $10.1198 \pm 5.8680$ & $0.4098 \pm 0.0040$ & $3.8393 \pm 0.4601$ & $0.0010 \pm 4.24\times10^{-5}$ \\
Spring--mass & ${7.3425}^{\dagger}$ & ${2.0733}^{\dagger}$ & ${4.9880}^{\dagger}$ & ${9.81\times10^{-5}}^{\dagger}$ \\
Double pendulum & ${36.2834}^{\dagger}$ & ${6.2625}^{\dagger}$ & ${5.0567}^{\dagger}$ & ${0.1434}^{\dagger}$ \\
\bottomrule
\end{tabular}
\end{table}

%% file: paper_sections/generated/visual_coupling.tex
\begin{table}[!htbp]
\caption{Complete nominal coupling comparison. A and B latent predictions are identical. Position RMSE units and seeds follow Table~\ref{tab:visual_main}.}
\label{tab:visual_coupling}
\centering
\setlength{\tabcolsep}{2pt}
\begin{tabular}{lrrrr}
\toprule
System & A latent & C latent & B structured & C structured \\
\midrule
Simple pendulum & $0.8814 \pm 0.1892$ & $0.9495 \pm 0.1089$ & $0.2212 \pm 0.0492$ & $0.2250 \pm 0.0445$ \\
Damped pendulum & $1.1261 \pm 0.0174$ & $1.0968 \pm 0.2974$ & $0.0446 \pm 0.0005$ & $0.0449 \pm 0.0003$ \\
Spring--mass & ${3.1486}^{\dagger}$ & ${3.2259}^{\dagger}$ & ${0.6905}^{\dagger}$ & ${0.6879}^{\dagger}$ \\
Double pendulum, arm 1 & ${0.5709}^{\dagger}$ & ${0.5621}^{\dagger}$ & ${0.1882}^{\dagger}$ & ${0.1782}^{\dagger}$ \\
\bottomrule
\end{tabular}
\end{table}

%% file: paper_sections/generated/planetary_motion_vectors.tex
\begin{table}[!htbp]
\caption{Eight-frame displacement vectors from autonomous forecasts, all 16 identity systems, mean $\pm$ sample SD across three seeds. Lower vector NMSE and higher cosine are better. Cosine excludes zero-length vectors, with valid coverage shown. NMSE includes all vectors. These quantities describe displacement over eight frames.}
\label{tab:planetary_motion_vectors}
\centering
\setlength{\tabcolsep}{2pt}
\begin{tabular}{lrrr}
\toprule
Model & Vector NMSE & Direction cosine & Coverage \\
\midrule
Absolute NTP & $1.8469 \pm 3.1033$ & $0.9456 \pm 0.0175$ & 81.3\% \\
Absolute + angular momentum & $0.0648 \pm 0.0250$ & $0.9516 \pm 0.0116$ & 79.1\% \\
Relative NTP & $0.1155 \pm 0.0599$ & $0.9617 \pm 0.0072$ & 86.5\% \\
Relative + TDV & $0.1081 \pm 0.0559$ & $0.9633 \pm 0.0020$ & 90.6\% \\
Pairwise NTP & $0.0301 \pm 0.0148$ & $0.9673 \pm 0.0044$ & 89.7\% \\
Pairwise + TDV & $0.0897 \pm 0.0321$ & $0.9660 \pm 0.0076$ & 86.7\% \\
\bottomrule
\end{tabular}
\end{table}

%% file: paper_sections/generated/planetary_full.tex
\begin{table}[!htbp]
\caption{All planetary objective conditions on the original-distribution test set. Single-seed conditions ($\dagger$) are exploratory. Plain NTP conditions containing TDV retain their original auxiliary coordinate encoding.}
\label{tab:planetary_full}
\centering
\setlength{\tabcolsep}{4pt}
\begin{tabular}{lrr}
\toprule
Model and supervision & Seeds & Position MSE (AU$^2$) \\
\midrule
Plain NTP + angular momentum & 3 & $0.1437 \pm 0.0741$ \\
Plain NTP + angular momentum + TDV & 3 & $0.1240 \pm 0.0448$ \\
Plain NTP + force supervision & 3 & $4.8811 \pm 6.2799$ \\
Plain NTP + force supervision + TDV & 3 & $3.1857 \pm 4.6176$ \\
Plain NTP + force-law supervision & 1 & ${0.4366}^{\dagger}$ \\
Plain NTP + force-law supervision + TDV & 1 & ${0.1326}^{\dagger}$ \\
Plain NTP + energy supervision & 1 & ${0.4465}^{\dagger}$ \\
Plain NTP + energy supervision + TDV & 1 & ${0.0898}^{\dagger}$ \\
Plain NTP & 3 & $1.0755 \pm 1.6393$ \\
Plain NTP + TDV & 3 & $0.3851 \pm 0.3196$ \\
Pairwise NTP + angular momentum & 1 & ${0.0568}^{\dagger}$ \\
Pairwise NTP + angular momentum + TDV & 1 & ${0.0984}^{\dagger}$ \\
Pairwise NTP & 3 & $0.0630 \pm 0.0292$ \\
Pairwise NTP + TDV & 3 & $0.1439 \pm 0.0444$ \\
Relative NTP & 3 & $0.2034 \pm 0.0864$ \\
Relative NTP + TDV & 3 & $0.1395 \pm 0.0542$ \\
\bottomrule
\end{tabular}
\end{table}

%% file: paper_sections/generated/angular_momentum.tex
\begin{table}[!htbp]
\caption{Normalized angular-momentum drift on the same 16 identity systems and 128-step forecasts. Lower is better. Drift measures temporal variation from predicted initial angular momentum, normalized by a fixed physical scale from the true initial state. The final column reports the mean and sample SD over training seeds.}
\label{tab:angular_momentum}
\centering
\setlength{\tabcolsep}{4pt}
\begin{tabular}{lrrrr}
\toprule
Model & 1337 & 1338 & 1339 & Mean $\pm$ SD \\
\midrule
Absolute NTP & 1.6417 & 0.4882 & 0.6290 & $0.9196 \pm 0.6293$ \\
Absolute + angular momentum & 0.4102 & 0.4148 & 0.5734 & $0.4662 \pm 0.0929$ \\
\bottomrule
\end{tabular}
\end{table}

%% file: paper_sections/generated/matched_force_all_models.tex
\begin{table}[!htbp]
\caption{Complete Earth force-adaptation comparison, seed 1337 and cadence 0.0189 years. Each cell reports direction cosine / physical vector NMSE on 424 held-out labels. Higher cosine and lower NMSE are better. Every condition is fitted separately.}
\label{tab:matched_force_all_models}
\centering
\setlength{\tabcolsep}{4pt}
\begin{tabular}{lrrr}
\toprule
Pretraining & Reference & $e=0.45$ & $2M_\star$ \\
\midrule
Plain NTP & 0.444 / 0.930 & -0.234 / 1.192 & 0.614 / 0.635 \\
NTP + force supervision & 0.211 / 1.202 & 0.467 / 0.968 & 0.222 / 1.362 \\
NTP + force law & -0.042 / 1.741 & 0.301 / 1.246 & 0.198 / 1.241 \\
NTP + angular momentum & 0.490 / 0.741 & 0.399 / 1.238 & 0.445 / 0.820 \\
NTP + TDV & 0.384 / 0.905 & 0.530 / 0.790 & 0.309 / 1.021 \\
Relative NTP & 0.560 / 0.759 & 0.263 / 0.786 & 0.400 / 0.939 \\
Relative + TDV & 0.858 / 0.282 & 0.776 / 0.489 & 0.786 / 0.354 \\
Pairwise NTP & 0.422 / 0.911 & 0.436 / 0.970 & 0.676 / 0.582 \\
Pairwise + TDV & 0.561 / 0.827 & 0.797 / 0.783 & 0.651 / 0.626 \\
\bottomrule
\end{tabular}
\end{table}

%% file: paper_sections/generated/matched_force_catalog.tex
\begin{table}[!htbp]
\caption{Paired improvements over plain NTP across eight named planets, two cadences, and three separately fitted conditions. These are related diagnostic cases from one pretrained seed, not independent replications.}
\label{tab:matched_force_catalog}
\centering
\setlength{\tabcolsep}{4pt}
\begin{tabular}{lrr}
\toprule
Pretraining & Direction wins & Vector-error wins \\
\midrule
NTP + force supervision & 21/48 & 23/48 \\
NTP + force law & 19/48 & 18/48 \\
NTP + angular momentum & 25/48 & 31/48 \\
NTP + TDV & 25/48 & 24/48 \\
Relative NTP & 33/48 & 23/48 \\
Relative + TDV & 34/48 & 36/48 \\
Pairwise NTP & 28/48 & 26/48 \\
Pairwise + TDV & 31/48 & 18/48 \\
\bottomrule
\end{tabular}
\end{table}

%% file: paper_sections/generated/force_alignment_transfer.tex
\begin{table}[!htbp]
\caption{Force-direction cosine after nominal force adaptation, higher-is-better. Values are mean $\pm$ sample SD across all 16 orbit/cadence cases, with 424 held-out labels per orbit and pretraining seed 1337. The same nominal-adapted model is evaluated in every column without refitting. The spread describes orbit variation, not independent training seeds.}
\label{tab:force_alignment_transfer}
\centering
\begin{tabular}{lrrr}
\toprule
Model & Reference & Eccentricity 0.45 & Stellar mass $\times 2$ \\
\midrule
Plain NTP & $0.3370 \pm 0.3007$ & $0.3081 \pm 0.2393$ & $0.2710 \pm 0.2163$ \\
NTP + angular momentum & $0.3627 \pm 0.2732$ & $0.2228 \pm 0.3047$ & $0.3470 \pm 0.2170$ \\
NTP + force supervision & $0.2802 \pm 0.4269$ & $0.1508 \pm 0.4302$ & $0.2153 \pm 0.3765$ \\
NTP + force law & $0.0938 \pm 0.3962$ & $0.1162 \pm 0.4126$ & $0.0633 \pm 0.3850$ \\
NTP + TDV & $0.2685 \pm 0.3399$ & $0.1524 \pm 0.2876$ & $0.2109 \pm 0.3068$ \\
Relative NTP & $0.3728 \pm 0.4578$ & $0.2846 \pm 0.4595$ & $0.2929 \pm 0.4164$ \\
Relative + TDV & $0.5441 \pm 0.3112$ & $0.5081 \pm 0.2957$ & $0.4128 \pm 0.2990$ \\
Pairwise NTP & $0.2462 \pm 0.4159$ & $0.3412 \pm 0.2851$ & $0.2352 \pm 0.3074$ \\
Pairwise + TDV & $0.2854 \pm 0.4193$ & $0.3702 \pm 0.2717$ & $0.2560 \pm 0.3403$ \\
\bottomrule
\end{tabular}
\end{table}

%% file: paper_sections/generated/planetary_system_intervals.tex
\begin{table}[!htbp]
\caption{Paired position-MSE differences in AU$^2$ on the original-distribution test set. Each system's difference is first averaged over the three fixed training seeds. Intervals use 10,000 paired bootstrap resamples of the 16 complete systems. Negative differences favor the first model. These intervals describe evaluation-system variation conditional on the trained checkpoints.}
\label{tab:planetary_system_intervals}
\centering
\setlength{\tabcolsep}{4pt}
\begin{tabular}{lrr}
\toprule
Contrast & Mean difference & 95\% interval \\
\midrule
Angular momentum - absolute NTP & -0.9318 & [-2.7508, 0.0248] \\
Pairwise - relative NTP & -0.1403 & [-0.3914, -0.0027] \\
Relative + TDV - relative NTP & -0.0639 & [-0.1836, 0.0006] \\
Pairwise + TDV - pairwise NTP & 0.0808 & [0.0058, 0.1946] \\
\bottomrule
\end{tabular}
\end{table}

%% file: paper_sections/generated/planetary_matched.tex
\begin{longtable}{>{\raggedright\arraybackslash}p{.36\linewidth}rrrr}
\caption{Complete matched scenario matrix for four models with matched coordinate encoding and three training seeds. Entries are mean autonomous planet-only MSE in AU$^2$. Each scenario is evaluated separately at its specified horizon.}\label{tab:planetary_matched}\\
\toprule
Scenario & Relative & + TDV & Pairwise & + TDV \\
\midrule
\endfirsthead
\toprule
Scenario & Relative & + TDV & Pairwise & + TDV \\
\midrule
\endhead
High eccentricity & 1.6329 & 0.7586 & 1.0505 & 1.3436 \\
Moderate eccentricity & 0.7782 & 0.2363 & 0.5053 & 0.6298 \\
400-frame horizon & 16.6076 & 4.9973 & 6.3577 & 15.3980 \\
Higher mass & 0.4404 & 0.2772 & 0.2251 & 0.5579 \\
Lower mass & 0.0988 & 0.0356 & 0.0618 & 0.0812 \\
Four planets & 104.3827 & 103.6287 & 50222.0733 & 64140.1797 \\
Four planets, reversed order & 111.1045 & 101.8626 & 52268.0930 & 71078.8401 \\
Matched reference & 0.2648 & 0.0922 & 0.2253 & 0.2007 \\
\bottomrule
\end{longtable}

%% file: paper_sections/generated/planetary_standard.tex
\begin{longtable}{>{\raggedright\arraybackslash}p{.36\linewidth}rrrr}
\caption{Complete standard scenario matrix for four models with matched coordinate encoding and three training seeds. Entries are mean autonomous planet-only MSE in AU$^2$. Each scenario is evaluated separately at its specified horizon.}\label{tab:planetary_standard}\\
\toprule
Scenario & Relative & + TDV & Pairwise & + TDV \\
\midrule
\endfirsthead
\toprule
Scenario & Relative & + TDV & Pairwise & + TDV \\
\midrule
\endhead
Broad eccentricity extrapolation & 67.4016 & 72.7884 & 66.9587 & 72.6492 \\
Broad eccentricity interpolation & 66.7530 & 69.5950 & 15219.3458 & 67.4005 \\
New nominal cohort & 119.4595 & 120.2306 & 99.9035 & 124.1985 \\
Six planets & 104.3705 & 106.7289 & 84825.8963 & 88123.6396 \\
Historical-distribution reference & 0.2034 & 0.1395 & 0.0630 & 0.1439 \\
Higher stellar-mass extrapolation & 119.7096 & 123.2674 & 12041.7800 & 16215.2600 \\
Lower stellar-mass extrapolation & 40.5245 & 49.6876 & 5007.6794 & 32.8092 \\
Four planets & 104.3827 & 103.6287 & 50222.0733 & 64140.1797 \\
Reflection & 1.3358 & 1.1612 & 1.7078 & 1.1384 \\
Rotation, 45 degrees & 0.2821 & 0.0622 & 0.0961 & 0.0832 \\
Shifted system distribution & 110.8523 & 106.8372 & 63675.5283 & 72313.6830 \\
Translation, (8, -6) & 0.2034 & 0.1395 & 0.0630 & 0.1439 \\
\bottomrule
\end{longtable}

%% file: paper_sections/generated/visual_shifts_simple_pendulum.tex
\begin{longtable}{>{\raggedright\arraybackslash}p{.36\linewidth}rr}
\caption{Simple pendulum: condition-B 400-step angle RMSE in radians. Mean and sample SD across available seeds, with one seed marked $\dagger$.}\label{tab:shifts_simple_pendulum}\\
\toprule
Scenario & Latent & Structured \\
\midrule
\endfirsthead
\toprule
Scenario & Latent & Structured \\
\midrule
\endhead
Changed background & $0.8802 \pm 0.1873$ & $0.2340 \pm 0.0671$ \\
Changed bob color & $0.8823 \pm 0.1872$ & $0.8999 \pm 0.1009$ \\
Gravity scaled by 0.5 & $0.9206 \pm 0.1782$ & $0.9942 \pm 0.0052$ \\
Gravity scaled by 1.5 & $0.8681 \pm 0.1911$ & $0.9087 \pm 0.0002$ \\
Higher initial velocity & $1.0800 \pm 0.1540$ & $0.9183 \pm 0.0922$ \\
Larger initial angle & $1.4310 \pm 0.1045$ & $1.3632 \pm 0.0929$ \\
Hidden length scaled by 0.5 & $0.8608 \pm 0.1925$ & $0.9230 \pm 0.0051$ \\
Hidden length scaled by 1.5 & $0.9002 \pm 0.1835$ & $0.9927 \pm 0.0054$ \\
Visible length scaled by 0.5 & $0.8650 \pm 0.1982$ & $0.9360 \pm 0.1636$ \\
Visible length scaled by 1.5 & $0.9042 \pm 0.1858$ & $0.9403 \pm 0.0322$ \\
Half-mass null control & $0.8814 \pm 0.1892$ & $0.2212 \pm 0.0492$ \\
Double-mass null control & $0.8814 \pm 0.1892$ & $0.2212 \pm 0.0492$ \\
Nominal & $0.8814 \pm 0.1892$ & $0.2212 \pm 0.0492$ \\
Translated pivot & $0.8850 \pm 0.1872$ & $0.9710 \pm 0.1187$ \\
Reflected scene & $0.8815 \pm 0.1891$ & $0.4203 \pm 0.0993$ \\
Zoom control & $0.8846 \pm 0.1904$ & $0.8145 \pm 0.0915$ \\
\bottomrule
\end{longtable}

%% file: paper_sections/generated/visual_shifts_real_pendulum.tex
\begin{longtable}{>{\raggedright\arraybackslash}p{.36\linewidth}rr}
\caption{Damped pendulum: condition-B 400-step angle RMSE in radians. Mean and sample SD across available seeds, with one seed marked $\dagger$.}\label{tab:shifts_real_pendulum}\\
\toprule
Scenario & Latent & Structured \\
\midrule
\endfirsthead
\toprule
Scenario & Latent & Structured \\
\midrule
\endhead
Changed background & $1.1246 \pm 0.0193$ & $0.0450 \pm 0.0004$ \\
Changed bob color & $1.1322 \pm 0.0122$ & $0.1658 \pm 0.0828$ \\
Gravity scaled by 0.5 & $1.1305 \pm 0.0191$ & $0.3701 \pm 0.0001$ \\
Gravity scaled by 1.5 & $1.1245 \pm 0.0172$ & $0.3274 \pm 0.0006$ \\
Higher initial velocity & $1.1708 \pm 0.0183$ & $0.1934 \pm 0.0099$ \\
Larger initial angle & $1.2401 \pm 0.0082$ & $0.5350 \pm 0.0237$ \\
Hidden length scaled by 0.5 & $1.1241 \pm 0.0169$ & $0.3516 \pm 0.0005$ \\
Hidden length scaled by 1.5 & $1.1281 \pm 0.0184$ & $0.3354 \pm 0.0002$ \\
Visible length scaled by 0.5 & $1.1220 \pm 0.0212$ & $0.3577 \pm 0.0543$ \\
Visible length scaled by 1.5 & $1.1258 \pm 0.0229$ & $0.3507 \pm 0.0037$ \\
Half mass, fixed drag & $1.1102 \pm 0.0184$ & $0.1105 \pm 0.0005$ \\
Double mass, fixed drag & $1.1547 \pm 0.0161$ & $0.1732 \pm 0.0006$ \\
Nominal & $1.1261 \pm 0.0174$ & $0.0446 \pm 0.0005$ \\
Translated pivot & $1.1319 \pm 0.0122$ & $0.4325 \pm 0.2594$ \\
Reflected scene & $1.1304 \pm 0.0122$ & $0.1048 \pm 0.0371$ \\
Zoom control & $1.1219 \pm 0.0239$ & $0.1256 \pm 0.0694$ \\
\bottomrule
\end{longtable}

%% file: paper_sections/generated/visual_shifts_spring_block.tex
\begin{longtable}{>{\raggedright\arraybackslash}p{.36\linewidth}rr}
\caption{Spring--mass: condition-B 400-step position RMSE in meters. Mean and sample SD across available seeds, with one seed marked $\dagger$.}\label{tab:shifts_spring_block}\\
\toprule
Scenario & Latent & Structured \\
\midrule
\endfirsthead
\toprule
Scenario & Latent & Structured \\
\midrule
\endhead
Changed background & ${3.1494}^{\dagger}$ & ${0.6923}^{\dagger}$ \\
Higher initial velocity & ${4.2494}^{\dagger}$ & ${1.9549}^{\dagger}$ \\
Higher mass & ${3.3534}^{\dagger}$ & ${2.7069}^{\dagger}$ \\
Lower mass & ${3.0442}^{\dagger}$ & ${2.1237}^{\dagger}$ \\
Mass/stiffness scaling control & ${3.1486}^{\dagger}$ & ${0.6905}^{\dagger}$ \\
Nominal & ${3.1486}^{\dagger}$ & ${0.6905}^{\dagger}$ \\
Higher stiffness & ${3.0442}^{\dagger}$ & ${2.1237}^{\dagger}$ \\
Lower stiffness & ${3.3534}^{\dagger}$ & ${2.7069}^{\dagger}$ \\
\bottomrule
\end{longtable}

%% file: paper_sections/generated/visual_shifts_double_pendulum.tex
\begin{longtable}{>{\raggedright\arraybackslash}p{.36\linewidth}rr}
\caption{Double pendulum, arm 1: condition-B 400-step angle RMSE in radians. Mean and sample SD across available seeds, with one seed marked $\dagger$.}\label{tab:shifts_double_pendulum}\\
\toprule
Scenario & Latent & Structured \\
\midrule
\endfirsthead
\toprule
Scenario & Latent & Structured \\
\midrule
\endhead
Higher gravity & ${0.5697}^{\dagger}$ & ${0.6518}^{\dagger}$ \\
Lower gravity & ${0.5691}^{\dagger}$ & ${0.6310}^{\dagger}$ \\
Longer first rod & ${0.5458}^{\dagger}$ & ${0.6366}^{\dagger}$ \\
Shorter first rod & ${0.6256}^{\dagger}$ & ${0.6873}^{\dagger}$ \\
Longer second rod & ${0.6035}^{\dagger}$ & ${0.6081}^{\dagger}$ \\
Shorter second rod & ${0.5451}^{\dagger}$ & ${0.5840}^{\dagger}$ \\
Common-mass scaling control & ${0.5709}^{\dagger}$ & ${0.1882}^{\dagger}$ \\
Higher initial velocity & ${0.7736}^{\dagger}$ & ${0.6343}^{\dagger}$ \\
Larger initial angles & ${1.0552}^{\dagger}$ & ${1.0248}^{\dagger}$ \\
Higher mass ratio & ${0.5904}^{\dagger}$ & ${0.5051}^{\dagger}$ \\
Lower mass ratio & ${0.5482}^{\dagger}$ & ${0.5013}^{\dagger}$ \\
Nominal & ${0.5709}^{\dagger}$ & ${0.1882}^{\dagger}$ \\
\bottomrule
\end{longtable}

%% file: paper_sections/13_revision_controls.tex
\begin{revision}
\section{Additional Matched Controls and Diagnostics}
\label{app:revision_controls}
\end{revision}

\begin{revision}
These experiments extend the submitted comparisons without replacing their
protocols or results. All predeclared seeds and valid unfavorable outcomes
are retained. A seed is an independently trained model, not an evaluation
episode. Means weight seeds equally and SD denotes sample standard deviation.
Paired wins describe consistency across seeds and are not significance tests.
Unless stated otherwise, visual controls use five seeds (42--46), 25 systems,
two observed images and 400 forecast steps. Planetary controls use five seeds
(1337--1341), 16 systems, 32 observed frames and 128 forecast frames.
Checkpoint selection uses training budget or validation data, never test scores.
\end{revision}

\begin{revision}
\subsection{Information and gradient paths}
\label{app:revision_information}
\end{revision}

\begin{revision}
\begin{table}[htbp]
\color{revisionblue}
\caption{Information supplied in each comparison. Additional inputs in the
equal-input planetary experiment are shared by both predictors.}
\centering
\begin{tabular}{p{.21\linewidth}p{.35\linewidth}p{.34\linewidth}}
\toprule
Comparison & Training information & Forecast information\\
\midrule
Visual observation model & Images and numerical positions/velocities & Initial images only\\
Energy/free, acceleration & Same simulated states and accelerations & Same decoded state and RK4\\
Energy/free, next state & Same simulated state transitions & Same decoded state and RK4\\
Energy/free, decoded & Same frozen-decoder transitions, no simulator targets & Same decoded state and RK4\\
Planetary scalar auxiliaries & Coordinates and targets computed from velocity/mass records & Coordinate prefix and cadence\\
Equal-input planetary & Same coordinates, velocity/mass inputs and modules, AM weight 0 or 0.1 & Same observed-prefix velocities and masses, no future velocity labels\\
Coefficient recovery & Frozen models and known nominal calibration & Observed fitting window, no test coefficient or future observations\\
\bottomrule
\end{tabular}
\end{table}
\end{revision}

\begin{revision}
Appendix~\ref{app:background_equations} and Figure~\ref{fig:prediction_paths}
define the visual prediction and gradient paths. The matched-head controls
freeze the observation model and vary only the transition, so any rollout
difference cannot come from a changed visual representation. The potential
and damping coefficient are learned, while the kinetic and dissipative forms
are specified. Planetary auxiliary losses update their readouts and the
backbone, whereas NTP gradients also train the adapter. The auxiliary
readouts are unused during generation.
\end{revision}

\begin{revision}
\subsection{Matched visual dynamics}
\label{app:revision_visual}
\end{revision}

\begin{revision}
Energy and free heads share examples, target information, shared-layer
initialization, optimizer budget, decoded initial states and RK4. Both use
$\dot q=v$. The free head predicts acceleration directly. The matched
acceleration objective omits the original energy-balance penalty. The
next-state objective supervises a one-step RK4 transition without acceleration
labels. Decoded-transition heads train on observation-model outputs without
simulator states or acceleration targets. Their decoder remains state-supervised.
No forecast-time optimizer updates or image resets occur.
\end{revision}

\begin{revision}
Simple-pendulum and spring heads have 132,609 parameters each. Damped heads
have 199,426 each. Double-pendulum energy/free heads have 133,121/133,378
parameters, a 0.19\% difference. Training uses 29,600 windows and 925 updates
per epoch, with 50/60/120 epochs for simple and spring/damped/double systems.
Final-epoch selection is fixed. Physical scores use the simulator's energy
function at the true evaluation parameters, not the learned potential.
For conservative single-coordinate systems we report energy drift. Damped
scores account for physical dissipation. Double-pendulum energy-change error
compares $E(t)-E(0)$ with the simulator's corresponding change.
\end{revision}

\begin{revision}
The simple-pendulum nominal angle benefit holds in every paired seed for all
three objectives. Damped and spring results distinguish consistency from
trajectory accuracy. Double-pendulum nominal angle reductions are small but
consistent for both links. Under next-state supervision, its free head has
lower physical energy-change error in every seed. Hidden-length angle errors
are nearly tied for the double pendulum. For decoded simple-pendulum heads,
the free head has lower hidden-length angle error in every seed despite higher
energy drift. Nominal physical consistency therefore does not establish
identification of an unseen law.
\end{revision}

\begin{revision}
\input{paper_sections/generated/revision_visual_full}
\input{paper_sections/generated/revision_visual_seeds}
\end{revision}

\begin{revision}
\subsection{Planetary capacity, supervision and geometry}
\label{app:revision_planetary}
\end{revision}

\begin{revision}
The generic relational adapter uses the same corrected relative-coordinate
encoding and embedding/logit entry points as geometry. Its message MLP takes
the four Cartesian components $[r_{ij},w_{ij}]$, uses a width-1536 GELU hidden
layer, and returns a width-768 embedding message and a two-component
displacement directly. Geometry instead takes the three invariant scalars
and predicts the two coefficients multiplying $r_{ij}$ and $w_{ij}$.
Both sum messages over source bodies and use the same zero-initialized gate
and Gaussian logit-bias interface (Appendix~\ref{app:equivariance}).
Added parameters are 1,191,171 for the generic adapter and 1,189,635 for geometry.
The five-seed,
twelve-angle mean position MSE is 0.2413 for generic, 0.1128 for geometric
and 0.1736 for Relative NTP. Geometry wins four paired seeds against generic
across rotations, versus two pairs on nominal forecasts. Separate branch
training gives the lowest nominal mean with both geometric branches and the
lowest 45-degree mean with embeddings only. The latter is a single-angle
comparison, not a ranking across all rotations.
\end{revision}

\begin{revision}
The nonconserved scalar target is $s\mu(r^{\mathsf T}v)$, with star-relative
position $r$, relative velocity $v$, and reduced mass $\mu$.
Its scale $s=\sqrt{\sum L_z^2/\sum[\mu(r^{\mathsf T}v)]^2}$ matches the
angular-momentum target's root mean square. Both sums use only valid
current-token target rows in the 999-row training windows, with no validation
or test data. The control retains the same scalar head, privileged state
records and training budget. AM beats Plain NTP in all five nominal pairs
and the radial target in three. High-error seeds remain in the mean and SD.
These results distinguish the chosen target from this specific auxiliary,
without identifying conservation as the unique cause of improvement.
\end{revision}

\begin{revision}
\begin{table}[htbp]
\color{revisionblue}
\caption{Position-only scalar-auxiliary control. Nominal position MSE in AU$^2$.}
\centering
\begin{tabular}{lrrr}
\toprule
Seed & Plain NTP & AM & Radial auxiliary\\
\midrule
1337 & 2.96764 & 0.21528 & 0.08199\\
1338 & 0.17692 & 0.14851 & 0.12026\\
1339 & 0.08195 & 0.06735 & 0.24541\\
1340 & 0.15014 & 0.12614 & 0.40307\\
1341 & 0.92143 & 0.12577 & 9.22451\\
\midrule
Mean & 0.85962 & 0.13661 & 2.01505\\
SD & 1.22697 & 0.05328 & 4.03216\\
Median & 0.17692 & 0.12614 & 0.24541\\
\bottomrule
\end{tabular}
\end{table}
\end{revision}

\begin{revision}
Equal-input training differs only in AM weight (0 or 0.1), apart from run
identifiers. Both predictors receive observed-prefix velocities and masses,
with matched modules, initialization and 15,625 updates. Future velocities
are never supplied. Mean nominal MSE falls by 40.2\% and median MSE by 28.4\%,
with three paired wins. Across the twelve conditions, AM has a lower mean in
three. Nominal relative angular-momentum drift is 0.6568/15.1345 and energy
drift 61.6551/9943.6186 without/with AM. Thus this matched-input nominal
trajectory gain does not imply improved conservation or uniform shift robustness.
\end{revision}

\begin{revision}
\setlength{\LTcapwidth}{\linewidth}
\setlength{\LTpre}{6pt}
\setlength{\LTpost}{6pt}
\input{paper_sections/generated/revision_planetary_tables}
\end{revision}

\begin{revision}
\subsection{Temporal-objective selection}
\label{app:revision_tdv}
\end{revision}

\begin{revision}
The complete weight grid uses validation NTP cross-entropy at seed 1337 and a
predeclared smaller-weight tie break. The four other seeds provide independent
confirmation. Existing endpoint models are reused rather than retrained.
Physical auxiliary weights remain 0.1, separate from the TDV grid.
\end{revision}

\begin{revision}
\begin{table}[H]
\color{revisionblue}
\caption{Complete TDV validation grid. Lower cross-entropy selects the weight.}
\centering
\begin{tabular}{lrrrrr}
\toprule
Representation & 0 & 0.01 & 0.03 & 0.1 & Selected\\
\midrule
Relative & 0.26604 & 0.28047 & 0.21167 & 0.21840 & 0.03\\
Pairwise & 0.24268 & 0.25905 & 0.29082 & 0.24178 & 0.1\\
\bottomrule
\end{tabular}
\end{table}
\end{revision}

\begin{revision}
Relative NTP has nominal MSE 0.1253 versus 0.2383 with selected TDV, with one
of four paired wins for TDV. Pairwise values are 0.0800/0.1459 with zero of
four wins. At fixed weight 0.1, the five-seed Relative comparison has three
nominal wins and four 45-degree wins. Pairwise TDV loses all five nominal
pairs while reducing mean 45-degree error by 44.0\%, with four wins.
Validation token loss did not select a consistently better nominal rollout
objective. The data establish the tradeoff, not its unique mechanism.
\end{revision}

\begin{revision}
\subsection{Unchanged-orbit intervention}
\label{app:revision_transfer}
\end{revision}

\begin{revision}
The original intervention uses 32 systems and three seeds. Adding or
reordering a noninteracting planet preserves the target's true orbit.
The original joint predictor nevertheless changes its forecast. Table~\ref{tab:revision_body}
retains all variants, including errors larger than persistence. Normalizing
messages did not resolve the failure. Factorized inference is a separate
follow-up: run the frozen predictor independently on each star-planet pair,
then combine predictions. Existing trajectories are bitwise unchanged under
addition and reordering. This procedure uses the benchmark's independent
central-force assumption, not learned interacting-body generalization.
\end{revision}

\begin{revision}
\begin{table}[htbp]
\color{revisionblue}
\caption{Original unchanged-orbit intervention, three-seed mean target-planet
MSE in AU$^2$. Each added planet is noninteracting.}
\label{tab:revision_body}
\centering
\begin{tabular}{lrrrr}
\toprule
Joint predictor & Original & Added $90^\circ$ & Added $180^\circ$ & Reordered\\
\midrule
Plain NTP & 5.542 & 197.042 & 198.688 & 352.059\\
Relative NTP & 0.662 & 5.040 & 5.046 & 5.109\\
Relative + TDV & 0.345 & 5.058 & 5.031 & 5.017\\
Pairwise NTP & 0.621 & 12794.888 & 4413.223 & 5358.292\\
Pairwise + TDV & 0.739 & 5.075 & 6438.685 & 5.090\\
Embedding only & 0.695 & 5.094 & 5.066 & 5.096\\
Logits only & 0.894 & 5.047 & 114.400 & 6.819\\
Generic adapter & 0.769 & 299.732 & 1050.744 & 854.773\\
Persistence & 5.075 & 5.075 & 5.075 & 5.075\\
Constant velocity & 10.848 & 10.848 & 10.848 & 10.848\\
\bottomrule
\end{tabular}
\end{table}
\end{revision}

\begin{revision}
On a separate 16-system confirmation cohort with three seeds, full-planet
MSE for one/two added planets is 1.3095/1.2659 for factorized Relative NTP,
1.0929/1.0754 for generic, and 0.5609/0.4888 for geometry, versus
6.0806/6.0806 for persistence. All factorized forecasts are finite.
These full-planet scores on a fresh cohort are not directly interchangeable
with the original target-only intervention scores.
\end{revision}

\begin{revision}
\subsection{Identifiable coefficient recovery}
\label{app:revision_recovery}
\end{revision}

\begin{revision}
We infer $g/L$ through a bounded acceleration multiplier, with known nominal
calibration $g/L=9.81$ and frozen neural weights. The test multipliers are
0.6, 0.8, 1.0, 1.2 and 1.4. Fitting uses a 32- or 64-frame window ending at
the common 64-frame observation cutoff, then forecasts 400 steps without
further observations. Adam optimizes only the log multiplier for 150 steps
at learning rate 0.05, bounded to $[0.25,2]$. Neither the test coefficient
nor future states enters fitting. There are 120 moving episodes and five
separately reported stationary controls per seed. Stationary contexts do not
identify a coefficient and are excluded from coefficient-error averages.
Gravity and length are not separately identifiable from this experiment,
and ideal-pendulum mass cancels.
\end{revision}

\begin{revision}
The 32-frame fit reduces image-derived energy-head forecast error by 16.8\%
in all five seeds. Free and classical dynamics yield similar image-derived
results. With true states the classical reference is most accurate, as
expected when the law and state are both known. The multiplier can compensate
observation or model error as well as physical change, so both coefficient
and forecast errors are reported. The longer fitting window is not uniformly
better. Five neural observation/head seeds do not constitute five independently
trained classical solvers.
\end{revision}

\begin{revision}
\input{paper_sections/generated/revision_recovery}
\end{revision}

\begin{revision}
\subsection{Observation correction and validation-gated retraining}
\label{app:revision_correction}
\end{revision}

\begin{revision}
A causal filter estimates final velocity from seven observed decoded
positions. It was selected on validation caches. Three seeds and fresh
confirmation conditions use the same 32-frame prefix, 400-step RK4 forecast
and frozen heads. For decoded simple-pendulum heads, initialization correction
changes energy/free angle RMSE from 0.34717/0.46618 to 0.25489/0.46280.
For decoded damped heads it changes 0.16327/0.16427 to 0.16092/0.16196.
All nominal contrasts improve in all three seeds, and all 4,608 nominal and
shifted forecasts are finite. Hidden-length decoded simple-pendulum error
still favors free dynamics (0.5203 versus 0.7349).
\end{revision}

\begin{revision}
The separate correction-based recovery experiment fits the last 32 states
from a 64-frame prefix. Mean relative coefficient error for energy/free/classical
dynamics changes from 21.183/20.819/21.185\% to 12.438/12.160/12.439\%.
Their adapted angle RMSE increases from 0.49060/0.49044/0.49048 to
0.55232/0.55300/0.55234. Better parameter estimation and better forecasting
remain distinct outcomes in this protocol.
\end{revision}

\begin{revision}
Retraining on corrected transitions follows a finite validation gate.
The simple-pendulum seed-42 pilot worsens validation mean angle RMSE from
0.39849 to 0.57178, so further seeds are skipped. Its confirmation error
also increases (0.37160 to 0.54092). The damped pilot passes validation,
allowing seeds 43--44. Table~\ref{tab:revision_correction} retains the complete
three-seed comparison. This changes training targets and initialization
together, separately from the initialization-only experiment above.
\end{revision}

\begin{revision}
\begin{table}[htbp]
\color{revisionblue}
\caption{Damped decoded-transition retraining, three-seed mean $\pm$ SD.
Raw/corrected training is paired with raw/corrected initialization.}
\label{tab:revision_correction}
\centering
\begin{tabular}{llrr}
\toprule
Condition / metric & Head & Raw & Corrected\\
\midrule
Nominal angle & Energy & $0.16376\pm0.03546$ & $0.10621\pm0.02165$\\
Nominal angle & Free & $0.16506\pm0.03601$ & $0.11188\pm0.02403$\\
Nominal energy balance & Energy & $2.48446\pm0.82061$ & $0.72888\pm0.26577$\\
Nominal energy balance & Free & $2.48845\pm0.81948$ & $0.69342\pm0.25715$\\
Hidden length, angle & Energy & 0.17662 & 0.23295\\
Hidden length, angle & Free & 0.17841 & 0.23567\\
Mass $\times2$, angle & Energy & 0.27968 & 0.23554\\
Mass $\times2$, angle & Free & 0.28073 & 0.24065\\
\bottomrule
\end{tabular}
\end{table}
\end{revision}

\begin{revision}
Nominal angle and balance errors improve in all three pairs for both heads.
Hidden-length forecasts worsen in every pair, whereas the fixed-drag mass
shift improves in every pair. All 1,536 forecasts are finite. These outcomes
were not used to retune the filter or extend the failed simple-pendulum pilot.
\end{revision}

\begin{revision}
\subsection{Direct image-based dynamics comparison}
\label{app:revision_named}

We evaluate the author Lag-caVAE architecture \citep{zhong2020lagcavae} on
simple-pendulum data across five seeds. Its 520,680 parameters train on
five-frame windows with the original reconstruction, spherical KL and 0.01
norm penalty. RGB images are averaged to grayscale and area-resized to
$32\times32$. Adam uses learning rate 0.001 without clipping or weight decay.
The native Euler solver and inferred coordinates are retained. Train-only
circular sign/offset calibration maps latent angles to physical units using
angle labels after training, without model updates. Our heads use a
state-supervised decoder and RK4. All methods observe two images. This
pipeline comparison complements matched-head attribution.
\end{revision}

\begin{revision}
Both checkpoint rules are fixed before testing: epoch 50 or image-validation
selection (500-epoch cap, minimum stopping epoch 100, checks every ten epochs,
patience of ten checks at 0.1\% relative improvement). Selected checkpoints
may precede epoch 100, and all runs stop below the cap. Coverage counts finite
400-step forecasts, not accuracy. Incomplete reports have undefined full-cohort
RMSE, never replaced by finite-only errors.
\end{revision}

\begin{revision}
\begin{table}[htbp]
\color{revisionblue}
\caption{Finite forecasts across five seeds and 25 systems per seed.
The denominator is 125 seed-system cases, not 125 independent systems.}
\centering
\begin{tabular}{lrrr}
\toprule
Method / checkpoint & Nominal & Hidden length $\times1.5$ & Mass $\times2$\\
\midrule
Lag-caVAE, epoch 50 & 37/125 & 39/125 & 37/125\\
Lag-caVAE, validation & 45/125 & 44/125 & 45/125\\
Matched energy, acceleration & 125/125 & 125/125 & 125/125\\
Matched free, acceleration & 125/125 & 125/125 & 125/125\\
\bottomrule
\end{tabular}
\end{table}
\end{revision}

\begin{revision}
\begin{table}[htbp]
\color{revisionblue}
\caption{Lag-caVAE stopping and selection. Coverage is nominal.}
\centering
\begin{tabular}{rrrrr}
\toprule
Seed & Stopped epoch & Selected epoch & Epoch-50 finite & Selected finite\\
\midrule
42 & 210 & 110 & 9/25 & 8/25\\
43 & 180 & 80 & 3/25 & 6/25\\
44 & 300 & 200 & 19/25 & 23/25\\
45 & 200 & 100 & 5/25 & 1/25\\
46 & 240 & 180 & 1/25 & 7/25\\
\bottomrule
\end{tabular}
\end{table}
\end{revision}

\begin{revision}
These failures describe Lag-caVAE's native protocol on this dataset, not a
general ranking of image-based mechanics. Both energy and free heads achieve
full coverage, so stability alone does not isolate energy structure.
\end{revision}

\begin{revision}
\subsection{Force-readout adaptation across pretraining seeds}
\label{app:revision_force}
\end{revision}

\begin{revision}
The post-hoc Earth illustration and its complete eight-planet, two-cadence
comparison use one pretraining seed (Appendices~\ref{app:matched_force_adaptation}
and~\ref{app:force_alignment_transfer}). Here five Plain/AM pretraining seeds
share fixed transfer-head and mask seeds, with 25 fitting, 50 validation and
424 held-out labels per orbit. Backbone and readout adapt jointly on nominal
labels, without shift refitting. Evaluation labels are held out on observed
orbits, not unseen autonomous trajectories.
\end{revision}

\begin{revision}
\begin{table}[htbp]
\color{revisionblue}
\caption{Five-seed force adaptation over all 16 orbit/cadence cases.
Direction cosine is higher-better and vector NMSE is lower-better.}
\centering
\begin{tabular}{lrrrrr}
\toprule
Condition & \multicolumn{2}{c}{Cosine} & AM wins & \multicolumn{2}{c}{Vector NMSE}\\
 & Plain & AM & & Plain & AM\\
\midrule
Nominal & 0.29422 & 0.28867 & 3/5 & 0.38040 & 0.46135\\
Eccentricity shift & 0.29609 & 0.27504 & 2/5 & 0.65567 & 0.68346\\
Stellar mass $\times2$ & 0.26615 & 0.26995 & 2/5 & 0.60695 & 0.64185\\
\bottomrule
\end{tabular}
\end{table}
\end{revision}

\begin{revision}
All reports have full direction coverage. AM has no consistent advantage in
this Plain/AM comparison, distinct from the original Relative+TDV contrast.
Forecasting, frozen probing and adapted readouts remain separate capabilities.
Vector NMSE of one denotes the zero-vector reference, not classification chance.
\end{revision}

%% file: paper_sections/generated/revision_visual_full.tex
\begin{revision}
\begin{longtable}{llrrr}
\caption{Simple, acceleration. Five-seed mean $\pm$ SD. Angle RMSE is in radians, position and physical energy RMSE use simulator units. Wins favor energy.}\\
\toprule
Condition & Metric & Energy & Free & Wins\\
\midrule\endfirsthead
\toprule
Condition & Metric & Energy & Free & Wins\\
\midrule\endhead
\bottomrule\endfoot
Nominal & Angle & $0.2147\pm0.03619$ & $0.2477\pm0.04156$ & 5/5\\
Nominal & Energy & $0.0001504\pm1.075e-05$ & $0.005856\pm0.00234$ & 5/5\\
Length $\times1.5$ & Angle & $0.9929\pm0.00383$ & $0.9915\pm0.004032$ & 0/5\\
Length $\times1.5$ & Energy & $1.803\pm0.006795$ & $1.803\pm0.006073$ & 3/5\\
Mass $\times2$ & Angle & $0.2147\pm0.03619$ & $0.2477\pm0.04156$ & 5/5\\
Mass $\times2$ & Energy & $0.0003008\pm2.15e-05$ & $0.01171\pm0.004681$ & 5/5\\
\end{longtable}

\begin{longtable}{llrrr}
\caption{Simple, next state. Five-seed mean $\pm$ SD. Angle RMSE is in radians, position and physical energy RMSE use simulator units. Wins favor energy.}\\
\toprule
Condition & Metric & Energy & Free & Wins\\
\midrule\endfirsthead
\toprule
Condition & Metric & Energy & Free & Wins\\
\midrule\endhead
\bottomrule\endfoot
Nominal & Angle & $0.2149\pm0.03618$ & $0.2574\pm0.03075$ & 5/5\\
Nominal & Energy & $0.0001453\pm1.257e-05$ & $0.006771\pm0.001653$ & 5/5\\
Length $\times1.5$ & Angle & $0.9929\pm0.00383$ & $0.991\pm0.003762$ & 0/5\\
Length $\times1.5$ & Energy & $1.803\pm0.006741$ & $1.8\pm0.008786$ & 1/5\\
Mass $\times2$ & Angle & $0.2149\pm0.03618$ & $0.2574\pm0.03075$ & 5/5\\
Mass $\times2$ & Energy & $0.0002907\pm2.515e-05$ & $0.01354\pm0.003307$ & 5/5\\
\end{longtable}

\begin{longtable}{llrrr}
\caption{Simple, decoded. Five-seed mean $\pm$ SD. Angle RMSE is in radians, position and physical energy RMSE use simulator units. Wins favor energy.}\\
\toprule
Condition & Metric & Energy & Free & Wins\\
\midrule\endfirsthead
\toprule
Condition & Metric & Energy & Free & Wins\\
\midrule\endhead
\bottomrule\endfoot
Nominal & Angle & $0.5117\pm0.02989$ & $0.6253\pm0.0028$ & 5/5\\
Nominal & Energy & $0.07246\pm0.005928$ & $4.274\pm0.04586$ & 5/5\\
Length $\times1.5$ & Angle & $0.9712\pm0.004439$ & $0.6832\pm0.0056$ & 0/5\\
Length $\times1.5$ & Energy & $1.976\pm0.01364$ & $6.866\pm0.07435$ & 5/5\\
Mass $\times2$ & Angle & $0.5117\pm0.02989$ & $0.6253\pm0.0028$ & 5/5\\
Mass $\times2$ & Energy & $0.1449\pm0.01186$ & $8.547\pm0.09172$ & 5/5\\
\end{longtable}

\begin{longtable}{llrrr}
\caption{Damped, acceleration. Five-seed mean $\pm$ SD. Angle RMSE is in radians, position and physical energy RMSE use simulator units. Wins favor energy.}\\
\toprule
Condition & Metric & Energy & Free & Wins\\
\midrule\endfirsthead
\toprule
Condition & Metric & Energy & Free & Wins\\
\midrule\endhead
\bottomrule\endfoot
Nominal & Angle & $0.04484\pm0.0005413$ & $0.04458\pm0.0006224$ & 0/5\\
Nominal & Energy & $0.0009714\pm5.04e-05$ & $0.005751\pm0.00437$ & 5/5\\
Length $\times1.5$ & Angle & $0.3355\pm0.0003184$ & $0.3352\pm0.0006161$ & 0/5\\
Length $\times1.5$ & Energy & $2.827\pm0.00366$ & $2.818\pm0.009271$ & 0/5\\
Mass $\times2$, fixed drag & Angle & $0.1733\pm0.0004713$ & $0.1735\pm0.0003907$ & 2/5\\
Mass $\times2$, fixed drag & Energy & $3.978\pm0.0194$ & $3.982\pm0.02138$ & 5/5\\
\end{longtable}

\begin{longtable}{llrrr}
\caption{Damped, next state. Five-seed mean $\pm$ SD. Angle RMSE is in radians, position and physical energy RMSE use simulator units. Wins favor energy.}\\
\toprule
Condition & Metric & Energy & Free & Wins\\
\midrule\endfirsthead
\toprule
Condition & Metric & Energy & Free & Wins\\
\midrule\endhead
\bottomrule\endfoot
Nominal & Angle & $0.04484\pm0.0005397$ & $0.04464\pm0.0006447$ & 1/5\\
Nominal & Energy & $0.0009968\pm4.941e-05$ & $0.005901\pm0.004818$ & 5/5\\
Length $\times1.5$ & Angle & $0.3355\pm0.000318$ & $0.3352\pm0.0006348$ & 0/5\\
Length $\times1.5$ & Energy & $2.827\pm0.003864$ & $2.818\pm0.01012$ & 0/5\\
Mass $\times2$, fixed drag & Angle & $0.1733\pm0.0004677$ & $0.1736\pm0.0003881$ & 2/5\\
Mass $\times2$, fixed drag & Energy & $3.978\pm0.0193$ & $3.982\pm0.02219$ & 5/5\\
\end{longtable}

\begin{longtable}{llrrr}
\caption{Damped, decoded. Five-seed mean $\pm$ SD. Angle RMSE is in radians, position and physical energy RMSE use simulator units. Wins favor energy.}\\
\toprule
Condition & Metric & Energy & Free & Wins\\
\midrule\endfirsthead
\toprule
Condition & Metric & Energy & Free & Wins\\
\midrule\endhead
\bottomrule\endfoot
Nominal & Angle & $0.2173\pm0.002325$ & $0.2188\pm0.00308$ & 5/5\\
Nominal & Energy & $3.865\pm0.03217$ & $3.872\pm0.04054$ & 3/5\\
Length $\times1.5$ & Angle & $0.2332\pm0.0007196$ & $0.2351\pm0.001137$ & 5/5\\
Length $\times1.5$ & Energy & $5.772\pm0.07476$ & $5.783\pm0.09535$ & 3/5\\
Mass $\times2$, fixed drag & Angle & $0.3306\pm0.001816$ & $0.3318\pm0.002419$ & 5/5\\
Mass $\times2$, fixed drag & Energy & $8.381\pm0.05029$ & $8.39\pm0.05694$ & 4/5\\
\end{longtable}

\begin{longtable}{llrrr}
\caption{Spring, acceleration. Five-seed mean $\pm$ SD. Angle RMSE is in radians, position and physical energy RMSE use simulator units. Wins favor energy.}\\
\toprule
Condition & Metric & Energy & Free & Wins\\
\midrule\endfirsthead
\toprule
Condition & Metric & Energy & Free & Wins\\
\midrule\endhead
\bottomrule\endfoot
Nominal & Position & $0.679\pm0.01534$ & $0.6789\pm0.01491$ & 2/5\\
Nominal & Energy & $0.0002241\pm0.0001551$ & $0.002225\pm0.001433$ & 5/5\\
Higher mass & Position & $2.719\pm0.008841$ & $2.719\pm0.008965$ & 2/5\\
Higher mass & Energy & $4.474\pm0.05802$ & $4.475\pm0.05747$ & 3/5\\
Higher stiffness & Position & $2.133\pm0.007501$ & $2.134\pm0.007409$ & 3/5\\
Higher stiffness & Energy & $4.459\pm0.05845$ & $4.458\pm0.05898$ & 1/5\\
\end{longtable}

\begin{longtable}{llrrr}
\caption{Spring, next state. Five-seed mean $\pm$ SD. Angle RMSE is in radians, position and physical energy RMSE use simulator units. Wins favor energy.}\\
\toprule
Condition & Metric & Energy & Free & Wins\\
\midrule\endfirsthead
\toprule
Condition & Metric & Energy & Free & Wins\\
\midrule\endhead
\bottomrule\endfoot
Nominal & Position & $0.679\pm0.01534$ & $0.6789\pm0.01492$ & 2/5\\
Nominal & Energy & $0.0002379\pm0.0001856$ & $0.001531\pm0.0008762$ & 4/5\\
Higher mass & Position & $2.719\pm0.008841$ & $2.719\pm0.009067$ & 1/5\\
Higher mass & Energy & $4.474\pm0.05803$ & $4.474\pm0.05914$ & 2/5\\
Higher stiffness & Position & $2.133\pm0.007501$ & $2.133\pm0.007452$ & 3/5\\
Higher stiffness & Energy & $4.459\pm0.05845$ & $4.459\pm0.05757$ & 3/5\\
\end{longtable}

\begin{longtable}{llrrr}
\caption{Double, acceleration. Five-seed mean $\pm$ SD. Angle RMSE is in radians, position and physical energy RMSE use simulator units. Wins favor energy.}\\
\toprule
Condition & Metric & Energy & Free & Wins\\
\midrule\endfirsthead
\toprule
Condition & Metric & Energy & Free & Wins\\
\midrule\endhead
\bottomrule\endfoot
Nominal & Upper angle & $0.1778\pm0.007065$ & $0.1795\pm0.007389$ & 5/5\\
Nominal & Lower angle & $0.2463\pm0.008734$ & $0.2487\pm0.00942$ & 5/5\\
Nominal & Energy & $0.1345\pm0.007913$ & $0.1375\pm0.007828$ & 4/5\\
Upper length $\times1.5$ & Upper angle & $0.634\pm0.001661$ & $0.6342\pm0.001155$ & 3/5\\
Upper length $\times1.5$ & Lower angle & $1.006\pm0.004361$ & $1.007\pm0.003565$ & 3/5\\
Upper length $\times1.5$ & Energy & $2.551\pm0.01789$ & $2.549\pm0.01307$ & 2/5\\
Lower mass $\times2$ & Upper angle & $0.5034\pm0.00999$ & $0.5033\pm0.01037$ & 2/5\\
Lower mass $\times2$ & Lower angle & $0.631\pm0.00976$ & $0.6309\pm0.01016$ & 2/5\\
Lower mass $\times2$ & Energy & $1.806\pm0.01441$ & $1.803\pm0.01702$ & 3/5\\
\end{longtable}

\begin{longtable}{llrrr}
\caption{Double, next state. Five-seed mean $\pm$ SD. Angle RMSE is in radians, position and physical energy RMSE use simulator units. Wins favor energy.}\\
\toprule
Condition & Metric & Energy & Free & Wins\\
\midrule\endfirsthead
\toprule
Condition & Metric & Energy & Free & Wins\\
\midrule\endhead
\bottomrule\endfoot
Nominal & Upper angle & $0.1777\pm0.007128$ & $0.1843\pm0.008724$ & 5/5\\
Nominal & Lower angle & $0.2462\pm0.008816$ & $0.2555\pm0.01144$ & 5/5\\
Nominal & Energy & $0.1338\pm0.008417$ & $0.08831\pm0.01137$ & 0/5\\
Upper length $\times1.5$ & Upper angle & $0.634\pm0.001654$ & $0.634\pm0.001459$ & 2/5\\
Upper length $\times1.5$ & Lower angle & $1.006\pm0.004314$ & $1.006\pm0.004348$ & 1/5\\
Upper length $\times1.5$ & Energy & $2.551\pm0.01809$ & $2.591\pm0.02467$ & 5/5\\
Lower mass $\times2$ & Upper angle & $0.5034\pm0.00998$ & $0.5089\pm0.01253$ & 5/5\\
Lower mass $\times2$ & Lower angle & $0.631\pm0.009739$ & $0.6362\pm0.01207$ & 5/5\\
Lower mass $\times2$ & Energy & $1.806\pm0.01459$ & $1.778\pm0.02065$ & 0/5\\
\end{longtable}
\end{revision}

%% file: paper_sections/generated/revision_visual_seeds.tex
\begin{revision}
\begin{longtable}{llrrrrr}
\caption{Nominal seed values for all matched visual controls. Columns identify training seeds. Angle RMSE is in radians, position and physical energy RMSE use simulator units. Energy/Free identify the transition heads.}\\
\toprule
Metric & Head & 42 & 43 & 44 & 45 & 46\\
\midrule\endfirsthead
\toprule
Metric & Head & 42 & 43 & 44 & 45 & 46\\
\midrule\endhead
\bottomrule\endfoot
\addlinespace
\multicolumn{7}{l}{Simple, acceleration}\\*
Angle & Energy & 0.2781 & 0.1968 & 0.1889 & 0.2093 & 0.2006\\*
Angle & Free & 0.3109 & 0.2399 & 0.1943 & 0.2445 & 0.2491\\*
Energy & Energy & 0.0001374 & 0.000161 & 0.0001617 & 0.0001434 & 0.0001484\\*
Energy & Free & 0.005858 & 0.006035 & 0.002667 & 0.009255 & 0.005464\\
\addlinespace
\multicolumn{7}{l}{Simple, next state}\\*
Angle & Energy & 0.2783 & 0.1968 & 0.1892 & 0.2093 & 0.2007\\*
Angle & Free & 0.312 & 0.2409 & 0.2395 & 0.245 & 0.2495\\*
Energy & Energy & 0.0001242 & 0.0001579 & 0.0001463 & 0.0001497 & 0.0001486\\*
Energy & Free & 0.006021 & 0.00919 & 0.005479 & 0.00776 & 0.005403\\
\addlinespace
\multicolumn{7}{l}{Simple, decoded}\\*
Angle & Energy & 0.5464 & 0.5055 & 0.5044 & 0.4691 & 0.5331\\*
Angle & Free & 0.6253 & 0.6231 & 0.6237 & 0.6301 & 0.6243\\*
Energy & Energy & 0.07946 & 0.06605 & 0.07498 & 0.06648 & 0.07534\\*
Energy & Free & 4.217 & 4.247 & 4.304 & 4.333 & 4.268\\
\addlinespace
\multicolumn{7}{l}{Damped, acceleration}\\*
Angle & Energy & 0.04482 & 0.04407 & 0.04498 & 0.04558 & 0.04475\\*
Angle & Free & 0.04482 & 0.04347 & 0.0449 & 0.04496 & 0.04473\\*
Energy & Energy & 0.0009265 & 0.0009985 & 0.001035 & 0.0009831 & 0.0009142\\*
Energy & Free & 0.002336 & 0.01199 & 0.003083 & 0.008721 & 0.002619\\
\addlinespace
\multicolumn{7}{l}{Damped, next state}\\*
Angle & Energy & 0.04481 & 0.04407 & 0.04498 & 0.04558 & 0.04475\\*
Angle & Free & 0.04482 & 0.04354 & 0.04489 & 0.04523 & 0.04472\\*
Energy & Energy & 0.001057 & 0.001005 & 0.0009945 & 0.001009 & 0.0009194\\*
Energy & Free & 0.001888 & 0.01356 & 0.003747 & 0.007633 & 0.002675\\
\addlinespace
\multicolumn{7}{l}{Damped, decoded}\\*
Angle & Energy & 0.218 & 0.2145 & 0.216 & 0.2207 & 0.2175\\*
Angle & Free & 0.2207 & 0.2145 & 0.2175 & 0.2226 & 0.2189\\*
Energy & Energy & 3.847 & 3.842 & 3.859 & 3.922 & 3.858\\*
Energy & Free & 3.862 & 3.836 & 3.863 & 3.942 & 3.857\\
\addlinespace
\multicolumn{7}{l}{Spring, acceleration}\\*
Position & Energy & 0.6906 & 0.6627 & 0.6936 & 0.6623 & 0.686\\*
Position & Free & 0.69 & 0.6629 & 0.6933 & 0.6628 & 0.6855\\*
Energy & Energy & 0.0001222 & 0.0001486 & 0.0002167 & 0.000139 & 0.0004941\\*
Energy & Free & 0.004296 & 0.00273 & 0.002327 & 0.001088 & 0.0006862\\
\addlinespace
\multicolumn{7}{l}{Spring, next state}\\*
Position & Energy & 0.6905 & 0.6628 & 0.6935 & 0.6621 & 0.6861\\*
Position & Free & 0.6899 & 0.6629 & 0.6933 & 0.6627 & 0.6855\\*
Energy & Energy & 0.0001554 & 0.0001557 & 0.0001231 & 0.0001879 & 0.0005674\\*
Energy & Free & 0.001756 & 0.002566 & 0.001094 & 0.001962 & 0.0002772\\
\addlinespace
\multicolumn{7}{l}{Double, acceleration}\\*
Upper angle & Energy & 0.1882 & 0.1697 & 0.181 & 0.1748 & 0.1753\\*
Upper angle & Free & 0.1904 & 0.171 & 0.1828 & 0.1775 & 0.1759\\*
Lower angle & Energy & 0.26 & 0.2387 & 0.2499 & 0.2406 & 0.2425\\*
Lower angle & Free & 0.2637 & 0.2402 & 0.252 & 0.2444 & 0.2433\\*
Energy & Energy & 0.1435 & 0.135 & 0.1386 & 0.133 & 0.1222\\*
Energy & Free & 0.1238 & 0.1429 & 0.1421 & 0.1384 & 0.1401\\
\addlinespace
\multicolumn{7}{l}{Double, next state}\\*
Upper angle & Energy & 0.1882 & 0.1696 & 0.1811 & 0.1746 & 0.175\\*
Upper angle & Free & 0.1916 & 0.1827 & 0.195 & 0.1764 & 0.1758\\*
Lower angle & Energy & 0.2599 & 0.2385 & 0.2499 & 0.2404 & 0.2423\\*
Lower angle & Free & 0.2665 & 0.2584 & 0.2655 & 0.2431 & 0.2437\\*
Energy & Energy & 0.1432 & 0.1347 & 0.1385 & 0.1318 & 0.1209\\*
Energy & Free & 0.0986 & 0.08305 & 0.07398 & 0.08477 & 0.1012\\
\end{longtable}
\end{revision}

%% file: paper_sections/generated/revision_planetary_tables.tex
\begin{revision}
\begin{longtable}{rrrr}
\caption{Complete twelve-angle sweep. Five-seed mean $\pm$ SD of position MSE in AU$^2$.}\\
\toprule
Angle (degrees) & Relative NTP & Generic & Geometric\\
\midrule\endfirsthead
\toprule
Angle (degrees) & Relative NTP & Generic & Geometric\\
\midrule\endhead
\bottomrule\endfoot
0 & $0.1635\pm0.08795$ & $0.2363\pm0.2426$ & $0.07358\pm0.026$\\
30 & $0.2618\pm0.1325$ & $0.2644\pm0.2701$ & $0.1694\pm0.1408$\\
60 & $0.07382\pm0.04821$ & $0.2818\pm0.3022$ & $0.04319\pm0.0272$\\
90 & $0.1615\pm0.07243$ & $0.213\pm0.1092$ & $0.1393\pm0.09102$\\
120 & $0.3822\pm0.1954$ & $0.2721\pm0.188$ & $0.1231\pm0.1183$\\
150 & $0.1115\pm0.07969$ & $0.2202\pm0.3057$ & $0.08296\pm0.0895$\\
180 & $0.1039\pm0.0598$ & $0.1692\pm0.1237$ & $0.2045\pm0.2118$\\
210 & $0.1745\pm0.1174$ & $0.2485\pm0.2293$ & $0.1557\pm0.09812$\\
240 & $0.06943\pm0.03597$ & $0.3098\pm0.3852$ & $0.07537\pm0.07456$\\
270 & $0.1802\pm0.09958$ & $0.2427\pm0.1711$ & $0.08175\pm0.04197$\\
300 & $0.2868\pm0.1689$ & $0.2968\pm0.3398$ & $0.1105\pm0.09752$\\
330 & $0.1144\pm0.08244$ & $0.1404\pm0.1228$ & $0.09409\pm0.07996$\\
\end{longtable}

\begin{longtable}{lrr}
\caption{Five-seed branch ablation, position MSE in AU$^2$.}\\
\toprule
Adapter & Nominal & Rotation $45^\circ$\\
\midrule\endfirsthead
\toprule
Adapter & Nominal & Rotation $45^\circ$\\
\midrule\endhead
\bottomrule\endfoot
Generic, both & $0.2348\pm0.2416$ & $0.2284\pm0.2339$\\*
Geometric, both & $0.07228\pm0.026$ & $0.1175\pm0.05299$\\*
Embedding only & $0.1798\pm0.1327$ & $0.07176\pm0.03799$\\*
Logits only & $0.1589\pm0.1269$ & $0.1114\pm0.09674$\\
\end{longtable}

\begin{longtable}{lrrr}
\caption{Individual branch-ablation seeds, position MSE in AU$^2$.}\\
\toprule
Adapter & Seed & Nominal & Rotation $45^\circ$\\
\midrule\endfirsthead
\toprule
Adapter & Seed & Nominal & Rotation $45^\circ$\\
\midrule\endhead
\bottomrule\endfoot
Generic, both & 1337 & 0.4787 & 0.6262\\
Generic, both & 1338 & 0.0469 & 0.2256\\
Generic, both & 1339 & 0.09138 & 0.08904\\
Generic, both & 1340 & 0.5175 & 0.1656\\
Generic, both & 1341 & 0.03945 & 0.03536\\
Geometric, both & 1337 & 0.04143 & 0.1422\\
Geometric, both & 1338 & 0.05145 & 0.06226\\
Geometric, both & 1339 & 0.09627 & 0.08382\\
Geometric, both & 1340 & 0.09953 & 0.1027\\
Geometric, both & 1341 & 0.07271 & 0.1963\\
Embedding only & 1337 & 0.09096 & 0.1025\\
Embedding only & 1338 & 0.06622 & 0.09245\\
Embedding only & 1339 & 0.3784 & 0.06633\\
Embedding only & 1340 & 0.2531 & 0.008062\\
Embedding only & 1341 & 0.1104 & 0.08947\\
Logits only & 1337 & 0.13 & 0.04685\\
Logits only & 1338 & 0.07981 & 0.1153\\
Logits only & 1339 & 0.1142 & 0.2777\\
Logits only & 1340 & 0.3829 & 0.05505\\
Logits only & 1341 & 0.08747 & 0.06218\\
\end{longtable}

\begin{longtable}{lrrr}
\caption{Complete equal-input comparison, five-seed mean $\pm$ SD of position MSE in AU$^2$.}\\
\toprule
Condition & AM weight 0 & AM weight 0.1 & AM wins\\
\midrule\endfirsthead
\toprule
Condition & AM weight 0 & AM weight 0.1 & AM wins\\
\midrule\endhead
\bottomrule\endfoot
Nominal & $0.2301\pm0.3155$ & $0.1377\pm0.1515$ & 3/5\\
Rotation $45^\circ$ & $0.07845\pm0.03767$ & $0.07003\pm0.04588$ & 3/5\\
Eccentricity interpolation & $188.9\pm27.18$ & $219.7\pm121.8$ & 3/5\\
Eccentricity extrapolation & $186.3\pm79.58$ & $217.9\pm57.93$ & 1/5\\
Fixed-ID cohort & $214.1\pm39.42$ & $284.4\pm100.9$ & 2/5\\
Low mass & $168\pm38.41$ & $183.1\pm119$ & 3/5\\
High mass & $155.4\pm47.19$ & $152.3\pm80.05$ & 3/5\\
Four bodies & $184\pm2.672$ & $187.8\pm8.43$ & 3/5\\
Six bodies & $180.1\pm0.9899$ & $181.4\pm2.577$ & 2/5\\
Reflection & $27.35\pm24.3$ & $47.92\pm50.75$ & 2/5\\
System shift & $334.9\pm16.04$ & $341.4\pm25.92$ & 3/5\\
Translation & $7.88\pm9.116$ & $35.8\pm47$ & 1/5\\
\end{longtable}

\begin{longtable}{lrr}
\caption{Individual equal-input nominal results, position MSE in AU$^2$.}\\
\toprule
Seed & AM weight 0 & AM weight 0.1\\
\midrule\endfirsthead
\toprule
Seed & AM weight 0 & AM weight 0.1\\
\midrule\endhead
\bottomrule\endfoot
1337 & 0.02623 & 0.1507\\*
1338 & 0.1314 & 0.3948\\*
1339 & 0.09797 & 0.07494\\*
1340 & 0.1047 & 0.03851\\*
1341 & 0.7903 & 0.02949\\*
Median & 0.1047 & 0.07494\\
\end{longtable}
\end{revision}

%% file: paper_sections/generated/revision_recovery.tex
\begin{revision}
\begin{longtable}{rlrrr}
\caption{Coefficient recovery from visual observations. Mean relative coefficient error and fixed/adapted angle RMSE (rad).}\\
\toprule
Window & Dynamics & Coefficient error & Fixed & Adapted\\
\midrule\endfirsthead
\toprule
Window & Dynamics & Coefficient error & Fixed & Adapted\\
\midrule\endhead
\bottomrule\endfoot
32 & Energy & 23.63243\% & 0.7475 & 0.6223\\
32 & Free & 23.57952\% & 0.7509 & 0.6237\\
32 & Classical & 23.63257\% & 0.7476 & 0.6223\\
64 & Energy & 25.19270\% & 0.7475 & 0.6701\\
64 & Free & 25.16573\% & 0.7509 & 0.6701\\
64 & Classical & 25.19806\% & 0.7476 & 0.67\\
\end{longtable}

\begin{longtable}{rlrrr}
\caption{Coefficient recovery from true state observations. Mean relative coefficient error and fixed/adapted angle RMSE (rad).}\\
\toprule
Window & Dynamics & Coefficient error & Fixed & Adapted\\
\midrule\endfirsthead
\toprule
Window & Dynamics & Coefficient error & Fixed & Adapted\\
\midrule\endhead
\bottomrule\endfoot
32 & Energy & 0.02996\% & 0.7203 & 0.001345\\
32 & Free & 0.38839\% & 0.7326 & 0.007654\\
32 & Classical & 0.00073\% & 0.72 & 7.223e-05\\
64 & Energy & 0.03036\% & 0.7203 & 0.0007468\\
64 & Free & 0.38652\% & 0.7326 & 0.005677\\
64 & Classical & 0.00073\% & 0.72 & 7.127e-05\\
\end{longtable}

\begin{longtable}{lrrrr}
\caption{Image-derived 32-frame recovery, individual observation/head seeds.}\\
\toprule
Dynamics & Seed & Coefficient error & Fixed & Adapted\\
\midrule\endfirsthead
\toprule
Dynamics & Seed & Coefficient error & Fixed & Adapted\\
\midrule\endhead
\bottomrule\endfoot
Energy & 42 & 28.272\% & 0.7614 & 0.6351\\
Energy & 43 & 20.620\% & 0.746 & 0.6389\\
Energy & 44 & 23.155\% & 0.7443 & 0.6009\\
Energy & 45 & 24.691\% & 0.7409 & 0.6039\\
Energy & 46 & 21.423\% & 0.745 & 0.6325\\
Free & 42 & 28.182\% & 0.7632 & 0.6363\\
Free & 43 & 20.552\% & 0.7508 & 0.6401\\
Free & 44 & 23.150\% & 0.7444 & 0.6006\\
Free & 45 & 24.633\% & 0.7458 & 0.606\\
Free & 46 & 21.381\% & 0.7504 & 0.6356\\
Classical & 42 & 28.270\% & 0.7619 & 0.6351\\
Classical & 43 & 20.621\% & 0.7459 & 0.6394\\
Classical & 44 & 23.155\% & 0.7443 & 0.6009\\
Classical & 45 & 24.697\% & 0.7409 & 0.6039\\
Classical & 46 & 21.419\% & 0.745 & 0.6324\\
\end{longtable}
\end{revision}